\pdfoutput=1
\documentclass[11pt]{article}

\usepackage[T1]{fontenc}
\usepackage[utf8]{inputenc}
\usepackage{times}
\usepackage{latexsym}
\usepackage{microtype}
\usepackage{inconsolata}

\usepackage{acl}

\usepackage{amsmath,amssymb,amsfonts}
\usepackage{booktabs}
\usepackage{multirow}
\usepackage{makecell}
\usepackage{graphicx}
\usepackage{subcaption}
\usepackage{xcolor}
\usepackage{algorithm}
\usepackage{algorithmic}

\DeclareMathOperator{\softplus}{softplus}
\DeclareMathOperator{\softmax}{softmax}

\title{Selective State-Space Adaptation and Retrieval \\for Language Model Reasoning}

\author{
Atahan Dokme \quad \quad Larry Heck \\
AI Virtual Assistant (AVA) Lab \\
Georgia Institute of Technology \\
\texttt{\{adokme3,larryheck\}@gatech.edu}
}
\begin{document}
\maketitle

\begin{abstract}
Low-rank adaptation introduces a static learned update applied identically to every input. The update provides task-level adaptation but does not explicitly represent token-level or instance-level state variation. A family of adapters is proposed that introduces selective state-space control at two complementary granularities. At the token level, \textbf{MaLoRA} (Mamba-modulated low-rank adaptation) makes the adapter's scaling factor a dynamic input-dependent function with recurrent state across tokens, in contrast to the stateless modulators of prior work. The token-level adapter improves over low-rank adaptation. On the other hand, it differentiates tokens by structural role but not by contextual relevance, which motivates placing evidence selection at the context level.  At the context level, \textbf{MaRA} (Mamba Retrieval Adapter) tracks cross-segment reasoning state and selects the segments most relevant to the query. State-space controlled retrieval of approximately three million parameters exceeds an eight-billion-parameter dense retriever on supporting-paragraph recall. Although base models perform poorly on the task without adaptation ($14$ to $25$ F1), MaRA recovers the evidence relevance latent in their representations. Across three frozen backbones and two multi-hop reasoning benchmarks, the end-to-end family improves reasoning accuracy on every cell of the 3-by-2 grid, by +6.4 F1 (+10.0\% relative) on average over the LoRA baseline.\footnote{Code available at \url{https://github.com/atahandokme/malora-mara}.}
\end{abstract}

\section{Introduction}
\label{sec:intro}

Low-rank adaptation \citep{hu2022lora} has become the standard approach for adjusting a frozen language model to a downstream task. The adapter is trained once and applied uniformly to every input. For tasks with multiple latent states, such as multi-step reasoning, this static update is shared across all of those states, encouraging reliance on training-set patterns rather than per-input adjustment. The goal of this study is to make adaptation more input-adaptive and better aligned with the state structure of the query, while leaving the backbone frozen.

Adaptation can operate at three levels: at the \emph{task} level, where LoRA already operates; at the \emph{token} level, where the strength of adaptation can become a function of the current token; and at the \emph{context} level, where the model can attend to the input segments most relevant to the query. The last two levels are nested: a context segment contains many tokens, so token-level control acts within the units that context-level control selects. Both mechanisms introduced here take the same form: a selective state-space module consumes a sequence derived from the frozen model's representations and emits a signal that governs a downstream quantity. At the token level that quantity is the magnitude of the low-rank update, applied during generation. At the context level it is the set of segments admitted to the input, decided before generation. The same Mamba block therefore plays two different jobs in the same system.

The first contribution of this work is \textbf{MaLoRA} (Mamba-modulated Low-Rank Adaptation), a recurrent token-level modulator for Low-Rank Adaptation. Unlike prior token modulators that use a stateless per-token scaling factor, MaLoRA uses a Mamba-based selective state-space module \citep{gu2024mamba} to propagate state across tokens.

Analysis of the trained modulator reveals a limit. Modulation is allocated by structural role and not by evidence relevance: supporting and distractor paragraphs receive mean values differing by at most $0.002$ in every cell (Section~\ref{sec:modulation-behavior}). Evidence selection therefore cannot be performed at the token level, whatever the modulator learns. The second contribution follows from this constraint. \textbf{MaRA} (Mamba Retrieval Adapter) applies the same state-space method one level up (from token level to context level), scoring each input segment from a state-space module over segment embeddings and selecting the most relevant segments for the adapted model.

The token-level modulation proves to be suppression-dominant and role-sensitive, allocating distinct values to marker, query and paragraph-content regions. At the segment level, evidence relevance proves recoverable from the frozen base model's representations even where the model answers the task poorly. The family is evaluated on MuSiQue \citep{trivedi-etal-2022-musique} and 2WikiMultihopQA \citep{ho-etal-2020-constructing} across three frozen backbones (Qwen-2.5-7B \citep{qwen2025qwen25technicalreport}, Llama-3.1-8B \citep{grattafiori2024llama3}, Gemma-2-9B \citep{riviere2024gemma2}). The combined system improves over the low-rank baseline on every cell of the $3{\times}2$ grid, and the gains from the two mechanisms largely accumulate: modulation reshapes the low-rank update by token role, while MaRA selects query-relevant segments before generation. RULER QA-2 \citep{hsieh2024rulerwhatsrealcontext} provides an out-of-distribution reasoning check at a context length at or beyond the adapters' training budget. Evidence for the token-level mechanism alone extends further, to commonsense and mathematical reasoning, while claims for the full system are scoped to supervised multi-hop question answering.
\vspace*{-.05in}

\section{Related Work}
\label{sec:related}

\paragraph{\mbox{Parameter-efficient adaptation.}}
Low-rank adaptation \citep{hu2022lora} adapts a frozen base model via a learned static update. Extensions add weight decomposition \citep{liu2024dora}, adaptive rank allocation \citep{zhang2023adaptive}, recurrent adapters \citep{nguyen2024readrecurrentadaptationlarge}, an asymmetric shared-$A$ multi-head architecture \citep{tian2024hydralora}, and mixture-of-LoRA routing \citep{dou-etal-2024-loramoe, gao-etal-2025-mola}. These approaches alter the structure of the low-rank update or route among several learned experts, but none place recurrent state across the tokens of a single input to control the magnitude of that update.

\paragraph{Token-aware adaptation.}
Making the low-rank update input-dependent at the token level has been explored recently. TopLoRA \citep{li2025beyond} learns a token-dependent \emph{diagonal} matrix between the LoRA factors and serves as the stateless prior-art baseline in this work. Neither this line nor the adapters above place recurrence across the tokens of a single input, which is the axis the token-level mechanism introduced here occupies.

\paragraph{State-space models in language modeling.}
State-space models provide sub-quadratic sequence modeling \citep{DBLP:journals/corr/abs-2111-00396, smith2023simplified, gu2024mamba, pmlr-v235-dao24a}. Two prior adapters apply such dynamics, and both place the recurrence \emph{across layers} of the adapter stack: \citet{nguyen2024readrecurrentadaptationlarge} (READ) propagates state through the adapter stack across layers, and \citet{yu-etal-2025-ssmlora} (SSMLoRA) uses an SSM to interconnect LoRA matrices inserted into the same module type at different layers. MaLoRA differs in \emph{axis}: a selective state-space module controls the magnitude of the LoRA update at each token, with recurrence \emph{across tokens of the same input}, so that each token's adapter strength carries memory of the trajectory leading up to it.

\paragraph{Reasoning and evidence selection.}
Multi-step QA benchmarks \citep{ho-etal-2020-constructing, trivedi-etal-2022-musique} expose a gap between the language model's possession of the answer and its use of the supporting evidence. Long-context evaluation surfaces related shifts \citep{hsieh2024rulerwhatsrealcontext, bai-etal-2024-longbench, laban2026llms}. A parallel line shows that information the model does not act on is nonetheless decodable from its hidden states, for instance signals of answer correctness recoverable by a linear probe \citep{orgad2024llmsknowmore}. Such probes have also been read for retrieval decisions, using intermediate hidden states to judge whether further documents are needed \citep{baek-etal-2025-probing}. Retrieval for multi-hop QA is otherwise handled by external modules, whether sparse \citep{Robertson2009ThePR} or dense and late-interaction \citep{karpukhin-etal-2020-dense, izacard2022unsuperviseddenseinformationretrieval, wang2024textembeddingsweaklysupervisedcontrastive, khattab2020colbert, zhang2025qwen3embeddingadvancingtext}, each trained separately from the model that consumes their output. Relatedly, adaptation objectives have been defined on representations rather than on outputs alone, aligning an adapted model's projected hidden states to a target latent subspace \citep{dokme2026temper}. The retrieval adapter introduced here is the case in which a decoded relevance signal is used to select evidence rather than only measured.


\section{MaLoRA: Stateful Token Modulation}
\label{sec:modulation}

\begin{figure*}[t]
  \centering
  \includegraphics[width=0.99\linewidth]{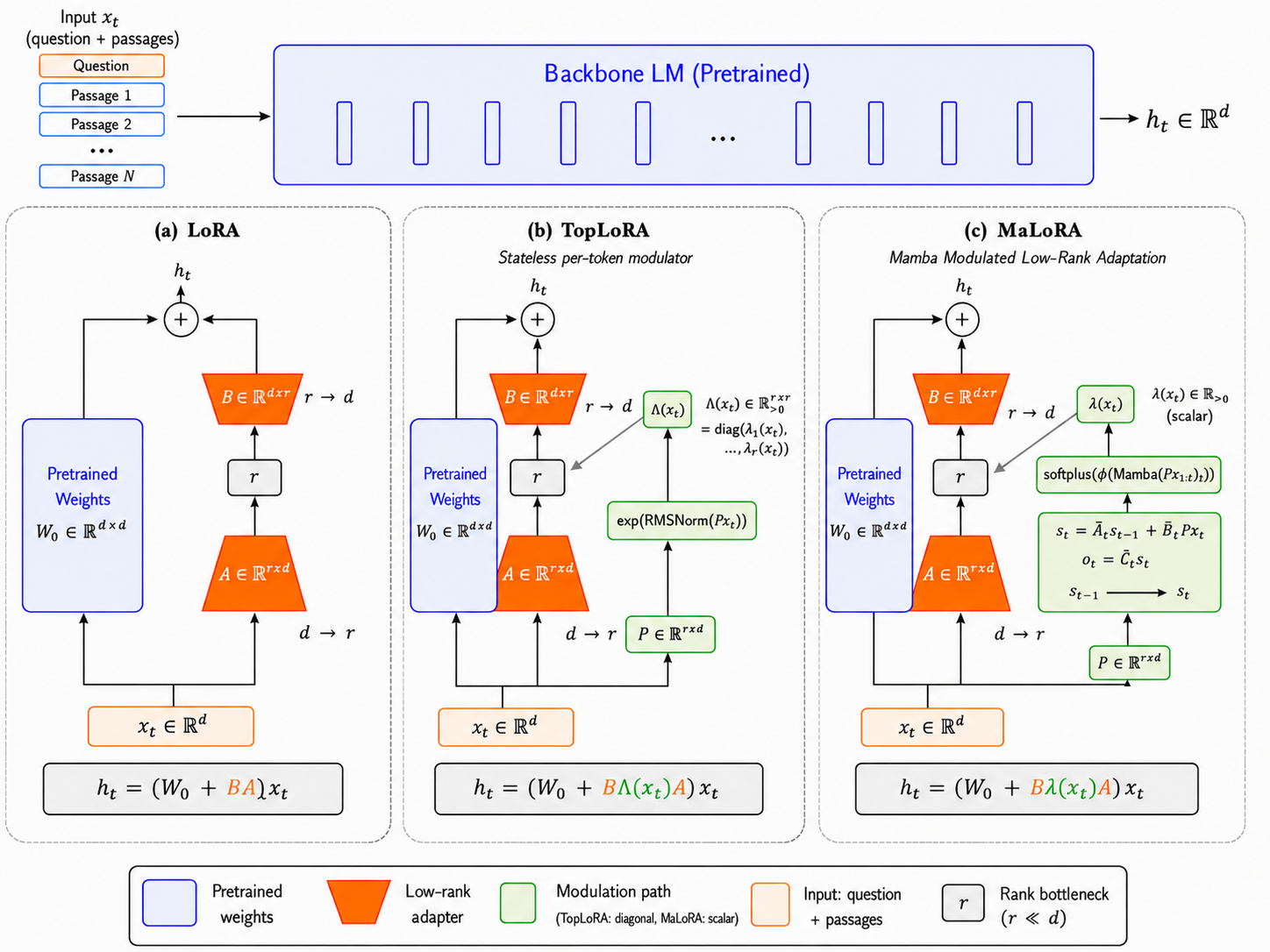}
  \caption{\textbf{From LoRA to MaLoRA.} \textbf{(a) LoRA}: a static low-rank update $BA x_t$ adds to the frozen output $W_0 x_t$. \textbf{(b) TopLoRA \citep{li2025beyond}}: a per-token diagonal modulator $\Lambda(x_t) = \mathrm{diag}\!\bigl(\exp(\mathrm{RMSNorm}(P x_t))\bigr)$ is inserted between the LoRA factors, yielding $h_t = (W_0 + B\,\Lambda(x_t)\,A)\,x_t$. The modulator is stateless: $\Lambda(x_t)$ depends only on the current token. \textbf{(c) MaLoRA}: the per-token head is replaced by a selective state-space module (Mamba) over the projected sequence $P x_{1:t}$, and the output is a scalar $\lambda(x_t) = \softplus(\phi(\mathrm{Mamba}(P x_{1:t})_t))$ that scales the $r$-dimensional intermediate. The modulator is stateful: $\lambda(x_t)$ depends on all positions up to $t$.}
  \label{fig:lora_lilora_malora}
\end{figure*}

\subsection{Preliminaries}
\label{sec:modulation-prelim}

\paragraph{Low-rank adaptation.}
Low-rank adaptation \citep{hu2022lora} reparameterises the update to a frozen weight $W_0 \in \mathbb{R}^{d_{\text{out}} \times d_{\text{in}}}$ as a low-rank product. For an input $x \in \mathbb{R}^{d_{\text{in}}}$, the output becomes
\begin{equation}
h \;=\; \bigl(W_0 \,+\, B A\bigr)\, x,
\quad A \in \mathbb{R}^{r \times d_{\text{in}}},\,
B \in \mathbb{R}^{d_{\text{out}} \times r},
\label{eq:lora}
\end{equation}
where $r \ll \min(d_{\text{in}}, d_{\text{out}})$ is the rank. The base weight $W_0$ is held frozen; only $A$ and $B$ are trained. The same update is applied to every token of every input, so the magnitude of the adaptation is independent of the model's representation of the current token.

\paragraph{TopLoRA: stateless per-token modulation.}
TopLoRA \citep{li2025beyond} relaxes this static assumption by inserting a learned per-token diagonal modulator $\Lambda(x_t) \in \mathbb{R}^{r\times r}_{>0}$ between the LoRA factors:
\begin{equation}
h_t \;=\; \bigl(W_0 \,+\, B\,\Lambda(x_t)\,A\bigr)\, x_t,
\label{eq:toplora}
\end{equation}
where $\Lambda(x_t) = \mathrm{diag}\bigl(\boldsymbol{\lambda}(x_t)\bigr)$ with $\boldsymbol{\lambda}(x_t) = \exp\!\bigl(\mathrm{RMSNorm}(P x_t)\bigr) \in \mathbb{R}^r_{>0}$ and $P \in \mathbb{R}^{r \times d_{\text{in}}}$ a learned low-rank projection (Fig.~\ref{fig:lora_lilora_malora}b). The modulator is \emph{stateless}: $\Lambda(x_t)$ depends only on the current token, with no propagation across positions.

\subsection{MaLoRA}
\label{sec:modulation-method}

MaLoRA replaces TopLoRA's stateless per-token diagonal with a stateful scalar modulator. The static low-rank update in Eq.~\ref{eq:lora} is multiplied by a per-token scalar $\lambda(x_t) \in \mathbb{R}_{>0}$ produced by a selective state-space module \citep{gu2024mamba} over the projected sequence:
\begin{equation}
\begin{aligned}
h_t \;&=\; \bigl(W_0 \,+\, B\,\lambda(x_t)\,A\bigr)\, x_t, \\
\lambda(x_t) \;&=\; \softplus\!\bigl(\phi(o_t)\bigr),
\end{aligned}
\label{eq:malora}
\end{equation}
where $P \in \mathbb{R}^{r \times d_{\text{in}}}$ is a learned low-rank projection sharing dimensions with $A$, $\phi$ is a small learned head, and $o_t$ is the Mamba output at position $t$. The Mamba module maintains a hidden state $s_t \in \mathbb{R}^{d_s}$ with input-dependent dynamics:
\begin{equation}
s_t \;=\; \bar{A}_t\, s_{t-1} \;+\; \bar{B}_t\, P x_t,
\quad o_t \;=\; \bar{C}_t\, s_t,
\label{eq:ssm}
\end{equation}
where $\bar{A}_t, \bar{B}_t, \bar{C}_t$ are computed from $P x_t$ in the standard Mamba parameterisation (Fig.~\ref{fig:lora_lilora_malora}c). Eq.~\ref{eq:malora} reduces to Eq.~\ref{eq:lora} when $\lambda(x_t) = 1$.

The Mamba recurrence block is the central contribution: it brings state inside the parameter-efficient adapter itself, giving the per-token modulator memory of the trajectory $x_1, \ldots, x_t$ rather than reacting only to $x_t$. The remaining design choices are as follows. \textbf{(i)} The output is a single scalar $\lambda(x_t)$ rather than TopLoRA's diagonal $\Lambda(x_t)$, which exposes one interpretable value per token for the behavioural analyses of Section~\ref{sec:modulation-behavior}. A diagonal output form was also trained and gives no consistent advantage (Appendix~\ref{app:ablations}). \textbf{(ii)} The activation is softplus so the modulator can both suppress ($\lambda < 1$) and amplify ($\lambda > 1$) the LoRA update; activation alternatives are ablated in Appendix~\ref{app:ablations}.

MaLoRA is trained jointly with the LoRA matrices, attached to $\{q\_\mathrm{proj},$ $k\_\mathrm{proj},$ $v\_\mathrm{proj},$ $\mathrm{up}\_\mathrm{proj},$ $\mathrm{down}\_\mathrm{proj}\}$ with projection- and layer-specific parameters. The Mamba module adds approximately $56\%$ to the LoRA parameter count, closely matching the $54\%$ added by the TopLoRA baseline. Exact counts are in Appendix~\ref{app:param_efficiency}.

\subsection{MaLoRA results}
\label{sec:modulation-results}

Two multi-hop QA datasets are used: MuSiQue \citep{trivedi-etal-2022-musique} ($2$--$4$ hops over $20$ candidates) and 2WikiMultihopQA \citep{ho-etal-2020-constructing} (up to $4$ hops over $10$ candidates). All three frozen backbones (Qwen-2.5-7B, Llama-3.1-8B, Gemma-2-9B) are evaluated at rank $r{=}16$ with EM/F1, beam $4$, over three seeds; full hyperparameters in Appendix~\ref{app:hparams}. MaLoRA improves over both LoRA and the TopLoRA baseline on every cell of Table~\ref{tab:modulation_main}, with the largest gains over LoRA on MuSiQue (up to $+7.9$ F1 on Llama). The gain over LoRA decomposes along two axes. Input-dependence alone, as in TopLoRA, lifts F1 across most cells (up to $+7.0$ F1 on MuSiQue Llama). Making that dependence stateful adds a further lift over TopLoRA on every cell. MaLoRA's advantage over the stateless modulator extends to commonsense reasoning (Appendix~\ref{app:commonsense}).

\begin{table}[t]
\centering
\footnotesize
\setlength{\tabcolsep}{1.5pt}
\renewcommand{\arraystretch}{1.25}
\caption{\textbf{Token modulators across three backbones.} F1 on MuSiQue and 2WikiMultihopQA, and RULER QA-2 accuracy (\%) at $4$k context using the MuSiQue-trained checkpoints. Means over three seeds; per-seed std in Appendix~\ref{app:ablations} (Table~\ref{tab:seed_std}).}
\label{tab:modulation_main}
\begin{tabular}{cl cc c}
\toprule
 & & MuSiQue & 2Wiki & QA-2 \\
\cmidrule(lr){3-3}\cmidrule(lr){4-4}\cmidrule(lr){5-5}
BB & Method & F1 & F1 & Acc@4k \\
\midrule
\multirow{4}{*}{\rotatebox[origin=c]{90}{Qwen-7B}}
 & Base                 & $14.4$ & $26.0$ & \textit{54.1} \\
 & LoRA \citep{hu2022lora}     & $51.1$ & $77.3$ & $56.8$ \\
 & TopLoRA \citep{li2025beyond} & $55.6$ & $77.9$ & $60.5$ \\
 & MaLoRA               & $\mathbf{56.4}$ & $\mathbf{79.3}$ & $\mathbf{60.7}$ \\
\midrule
\multirow{4}{*}{\rotatebox[origin=c]{90}{Llama-8B}}
 & Base                 & $17.4$ & $31.9$ & \textit{10.8} \\
 & LoRA \citep{hu2022lora}     & $54.4$ & $78.5$ & $60.9$ \\
 & TopLoRA \citep{li2025beyond} & $61.4$ & $80.4$ & $63.9$ \\
 & MaLoRA               & $\mathbf{62.3}$ & $\mathbf{83.3}$ & $\mathbf{64.7}$ \\
\midrule
\multirow{4}{*}{\rotatebox[origin=c]{90}{Gemma-9B}}
 & Base                 & $25.3$ & $34.2$ & \textit{38.2} \\
 & LoRA \citep{hu2022lora}     & $59.8$ & $77.2$ & $64.4$ \\
 & TopLoRA \citep{li2025beyond} & $59.4$ & $79.2$ & $64.6$ \\
 & MaLoRA               & $\mathbf{63.7}$ & $\mathbf{82.2}$ & $\mathbf{65.1}$ \\
\bottomrule
\end{tabular}
\end{table}
\subsection{Interpretations of MaLoRA Modulation}
\label{sec:modulation-behavior}

\begin{figure}[b]
  \centering
  \includegraphics[width=1.00\linewidth]{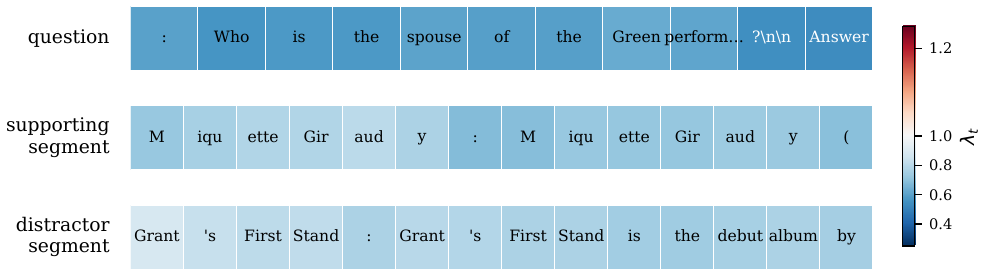}
  \caption{\textbf{Per-token modulation $\lambda(x_t)$ for MaLoRA on a MuSiQue example} (Llama-3.1-8B, seed-$43$). Cell colour averages $\lambda(x_t)$ over $160$ modulated projections; darker = stronger suppression. Question tokens are suppressed most; supporting and distractor content retain higher $\lambda$. Per-region breakdown in Table~\ref{tab:region_lambda}.}
  \label{fig:token_heatmap}
\end{figure}

The trained modulator is suppression-dominant: most $\lambda(x_t)$ fall below one, at $61$ to $84\%$ across the three cells examined, with structural separators and query tokens suppressed most heavily and content tokens retaining higher values (Fig.~\ref{fig:token_heatmap}). Modulation varies far more across tokens than across projections or layers, so the per-token component cannot be reproduced by any static rescaling. Suppression is deepest in $v\_\mathrm{proj}$ in every cell. The final transformer layer is suppressed on Qwen-2.5-7B and Llama-3.1-8B, but not on Gemma-2-9B (Appendix~\ref{app:proj_layer_marginals}).

\paragraph{Selective state-space usage.}
A direct check compares the Mamba block's stateful output $z_t = \mathrm{Mamba}(P x_{1:t})_t$ against a stateless counterfactual $z_t^{\mathrm{stateless}} = \mathrm{Mamba}(P x_t)$ that processes each position independently. The relative discrepancy $\|z_t - z_t^{\mathrm{stateless}}\| / \|z_t\|$, averaged across the six MaLoRA cells (Qwen/Llama/Gemma $\times$ MuSiQue/2WikiMultihopQA), rises from $0.13$ at position $1$ to $0.61$--$0.63$ across mid-prompt positions and remains in that range out to the longest positions tested. The stateless variant therefore diverges from the stateful one by a non-trivial fraction at every position past the first token: the SSM state is genuinely accumulating information rather than acting as a decorative recurrence. Per-cell curves are in Appendix~\ref{app:state_probe}. The selective state-space formulation was adopted because its input-dependent state transitions match the latent state structure of multi-hop reasoning. Comparisons against other recurrent families demonstrate that at the token level, GRU and LSTM modulators are not consistently separable from the Mamba modulator, so the operative ingredient there is recurrence itself. At the segment level, where each position summarises a full paragraph and the reasoning state contextually transitions, the selective module exceeds both classical recurrences by more than $20$ recall points (Section~\ref{sec:routing-ablation}, Appendix~\ref{app:recurrent}).

Aggregating $\lambda(x_t)$ over prompt regions on the full validation set across the three MaLoRA-backbone cells (Table~\ref{tab:region_lambda}; extended discussion in Appendix~\ref{app:region_lambda}) shows a systematic role-based pattern: \textbf{the modulator differentiates tokens by their structural role}. Marker, question and paragraph-content regions receive distinct mean values in every cell, and the separation is large. On Llama-3.1-8B, question tokens average $0.603$ against $0.708$ for paragraph content, a difference of roughly $15\%$ in how strongly the adapter is applied. Figure~\ref{fig:token_heatmap} shows the same effect within a single example, where the question span is the darkest region of the prompt. The direction is not uniform across backbones: Qwen and Llama suppress marker and query tokens relative to paragraph content, whereas Gemma amplifies them on the checkpoint analysed ($1.212$ and $1.730$). Within paragraph content, however, the separation vanishes. Supporting and distractor paragraphs receive mean values differing by at most $0.002$, two orders of magnitude below the role-based differences just described. Token-level modulation thus reshapes how strongly the adapter applies to different prompt roles, which is a useful axis of adaptation, but evidence selection requires a separate mechanism at the segment level (Section~\ref{sec:routing}).

\begin{table}[t]
  \centering\small
  \setlength{\tabcolsep}{3pt}
  \caption{\textbf{Mean $\lambda(x_t)$ by token region} (full MuSiQue validation, MaLoRA across three backbones). Mark.\ / Ques.\ / Supp.\ / Dist.\ = marker / question / supporting / distractor tokens. Discussion in Section~\ref{sec:modulation-behavior}.}
  \label{tab:region_lambda}
  \begin{tabular}{l cccc}
    \toprule
    BB & Mark.\ & Ques.\ & Supp.\ & Dist.\ \\
    \midrule
    Qwen  & $0.792$ & $0.772$ & $0.904$ & $0.904$ \\
    Llama & $0.725$ & $0.603$ & $0.708$ & $0.709$ \\
    Gemma & $1.212$ & $1.730$ & $0.929$ & $0.930$ \\
    \bottomrule
  \end{tabular}
\end{table}

\section{MaRA: Context-Level Adaptation}
\label{sec:routing}

MaRA operates one level above the token modulator: the input is split into context segments, each segment is scored by relevance to the query, and the top-$k$ are exposed to the modulated language model for generation. The adapter is small ($\approx 3$M trainable parameters) and introduces no separate retrieval encoder: it reads the frozen backbone's own intermediate hidden states and adds only a segment-level mixer, with the recurrence running over segment embeddings rather than tokens. In comparison to the token-level modulator it introduces no weight adaptation per token. Instead, it retrieves which segments the model needs for reasoning from its latent representations, using a state-space method. The evidence information already resides in the backbone's hidden states; the segment-level state-space module tracks the reasoning chain across segments and surfaces the supporting ones, in contrast to the token-level modulator's paragraph-level relevance-blindness (Section~\ref{sec:modulation-behavior}). Full per-stage exposition is in Appendix~\ref{app:router_details}.

\begin{figure*}[t]
  \centering
  \includegraphics[width=1.00\linewidth]{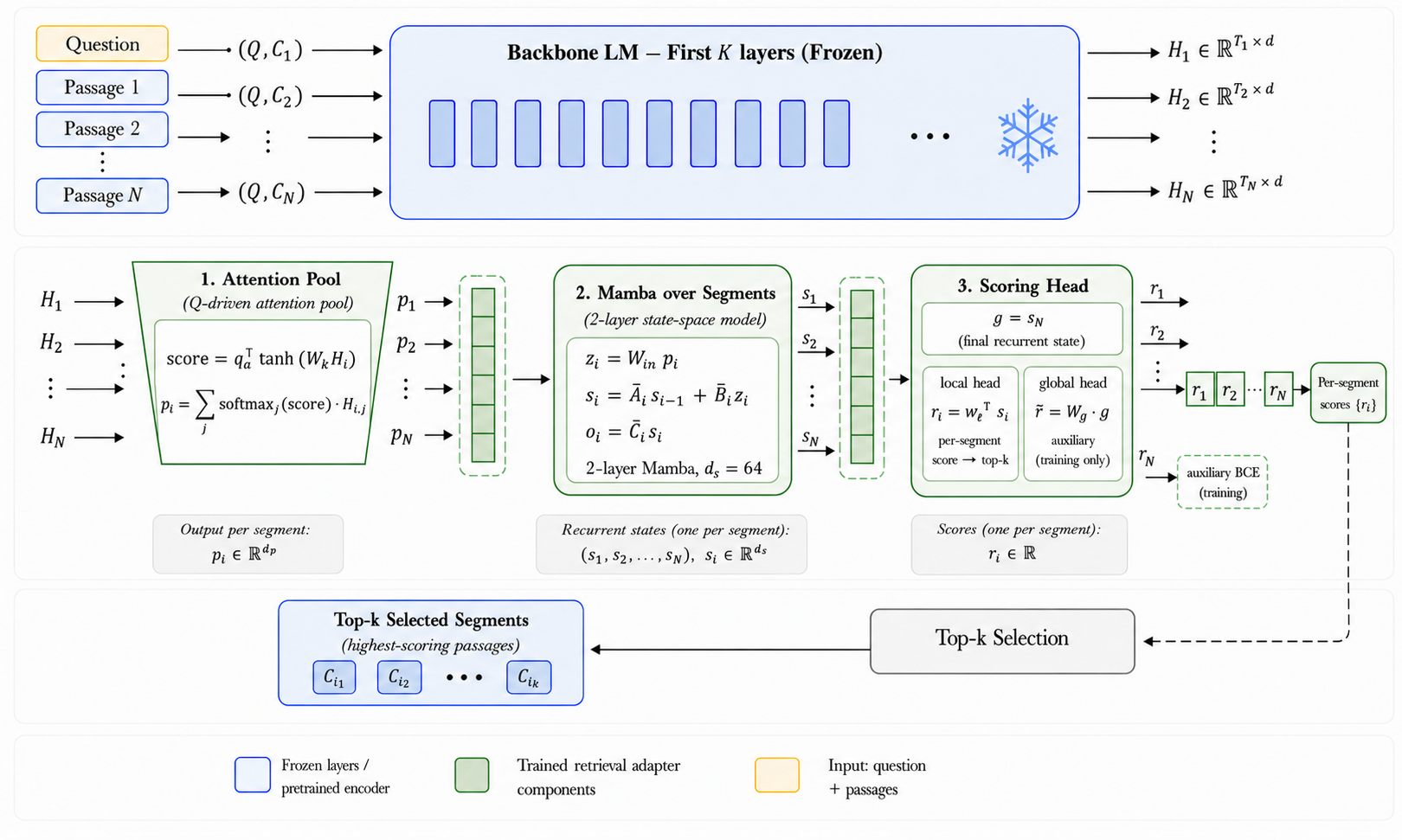}
  \caption{\textbf{MaRA architecture and data flow.} Each (query, segment) pair $(Q, C_i)$ is encoded by the first $K$ layers of the frozen backbone, yielding hidden states $H_i$. A learned attention pool reduces each $H_i$ to a single embedding $p_i$. A two-layer Mamba runs over the segment-level sequence $(p_1, \ldots, p_N)$, producing recurrent states $(s_1, \ldots, s_N)$. A local scoring head maps each recurrent state $s_i$ to a relevance score $r_i$; a separate global head reads the final state $g = s_N$ and provides auxiliary supervision during training.}
  \label{fig:router_arch}
\end{figure*}

\paragraph{Method.}
Given a query $Q$ and $N$ segments $\mathcal{C}=(C_1,\ldots,C_N)$, the retrieval adapter proceeds in four stages.

\emph{(i) Segment encoder.} Each $C_i$ is concatenated with $Q$ and passed through the first $K$ layers of the frozen backbone, producing hidden states $H_i \in \mathbb{R}^{T_i \times d}$. The depth $K$ is set per backbone: $16$ for Qwen, $12$ for Llama, and $20$ for Gemma on MuSiQue and $16$ on 2WikiMultihopQA (Appendix~\ref{app:hparams}). No new encoder parameters are introduced.

\emph{(ii) Attention pool.} A learned query vector $q_a$ attends over each segment's hidden states with an additive score, and the segment embedding is the attention-weighted sum of those hidden states. The hidden states are already query-conditioned, since each segment is encoded together with $Q$ in stage (i). With $W_k \in \mathbb{R}^{d_p \times d}$,
\begin{equation}
p_i = \sum_{j=1}^{T_i} \softmax_j\!\big(q_a^{\!\top}\tanh(W_k H_{i,j})\big) H_{i,j}.
\label{eq:pool}
\end{equation}

\emph{(iii) Mamba over segments.} The segment embeddings are passed through a two-layer selective state-space module \citep{gu2024mamba}, with the recurrence running over \emph{segments} (not tokens):
\begin{equation}
\begin{aligned}
z_i \;&=\; W_{\mathrm{in}}\, p_i, \\
(s_1, \ldots, s_N) \;&=\; \mathrm{Mamba}\!\big(z_1, \ldots, z_N\big),
\end{aligned}
\label{eq:mamba_seg}
\end{equation}
where $W_{\mathrm{in}}$ projects each pooled segment embedding to the recurrence width.
The recurrence is causal: $s_i$ depends on segments $1$ through $i$. The unit of state evolution is one context segment, matching the granularity at which contextual transitions occur in multi-step reasoning (Section~\ref{sec:modulation-behavior}).

\emph{(iv) Scoring heads.} Two linear heads read the recurrent states. A local head scores each segment from its own state $s_i$, and a global head reads the final state $g = s_N$ (which has integrated all $N$ segments through the causal recurrence) and predicts all $N$ labels jointly:
\begin{equation}
g \;=\; s_N,
\quad
r_i \;=\; \mathbf{w}_{\ell}^{\!\top}\, s_i,
\quad
\tilde{\mathbf{r}} \;=\; W_{g}\, g \;\in\; \mathbb{R}^{N}.
\label{eq:score}
\end{equation}
The per-segment score $r_i$ is used to rank and select the top-$k$ segments, which are forwarded to the modulator-equipped language model. The global head $\tilde{\mathbf{r}}$ supplies an auxiliary supervision signal that conditions the shared recurrence (Appendix~\ref{app:router_details}); it is not used at selection time.

\paragraph{Training objective.}
Binary supporting-segment labels $y_i \in \{0,1\}$ supervise the retrieval adapter. The loss combines weighted BCE on the local scores (positive class up-weighted to offset class imbalance), an intra-example pairwise margin that pushes supporting scores above non-supporting scores, and an auxiliary BCE on the global head:
\begin{equation}
\begin{aligned}
\mathcal{L} \;=\; & \sum_i \mathrm{BCE}(r_i, y_i) \\
            & + \beta \sum_{i:\, y_i=1}\sum_{j:\, y_j=0} \max(0,\, m - (r_i - r_j)) \\
            & + \gamma \sum_i \mathrm{BCE}(\tilde{r}_i, y_i),
\end{aligned}
\label{eq:loss-router}
\end{equation}
where $\tilde{r}_i$ is the $i$-th logit of the global head (Eq.~\ref{eq:score}) and $\gamma{=}0.5$. Selection uses only the local scores $r_i$; the global term is an auxiliary regularizer on the shared recurrence.
The retrieval adapter is trained against the unadapted backbone, with no modulator attached; no end-to-end fine-tuning of the two stages is performed. Hyperparameters and the canonical training pool are detailed in Appendix~\ref{app:hparams}.

\paragraph{Chain selection.}
At inference the retrieval adapter is applied over $N_p$ chained passes. Each pass re-scores the candidate segments conditioned on the segments already selected, then appends the highest-scoring unselected segment to an ordered anchor set. The final top-$k$ places these anchors first and fills the remainder by the last pass's scores. $N_p{=}4$ passes are used on both datasets, matching the pass budget the retrieval adapter was trained with. The procedure is given in Appendix~\ref{app:iter}.

\subsection{MaRA architecture comparison}
\label{sec:routing-ablation}

MaRA is compared to eight alternatives under matched supervision, on MuSiQue and 2WikiMultihopQA with the base language model (no adapter). All architectures are scored in a single pass, so the numbers here isolate the architecture rather than the chained procedure used at inference (Appendix~\ref{app:iter}). Each alternative retains MaRA's frozen encoder slice, attention pool and scoring heads, and replaces one component. The alternatives ablate the per-segment pool (MeanPool, LastTok), the segment-level mixer (Transformer, MLP, PoolOnly), the recurrent family (GRU, LSTM, two-layer classical recurrences at the same width as the Mamba mixer), and the recurrence granularity (TokenMamba, recurrence over concatenated tokens instead of pooled segments). See Appendix~\ref{app:router_details} for definitions.

\begin{table}[t]
\centering
\small
\setlength{\tabcolsep}{4pt}
\caption{\textbf{MaRA architecture comparison.} Recall@$4$ on MuSiQue and 2WikiMultihopQA validation splits, no adapter attached, single scoring pass. Best per column in bold. Trainable parameters (MuSiQue/Qwen): LastTok $3.6$k, PoolOnly $1.8$M, TokenMamba $2.0$M, GRU $2.6$M, Mamba $2.9$M, MeanPool $2.9$M, LSTM $2.9$M, Transformer $3.4$M, MLP $13.4$M.}
\label{tab:router_arch}
\begin{tabular}{lcccc}
\toprule
& \multicolumn{2}{c}{MuSiQue} & \multicolumn{2}{c}{2WikiMultihopQA} \\
\cmidrule(lr){2-3}\cmidrule(lr){4-5}
Architecture & Qwen & Llama & Qwen & Llama \\
\midrule
MLP            & 20.6 & 19.6 & 93.8 & 95.3 \\
LastTok        & 30.1 & 32.3 & 84.8 & 83.7 \\
TokenMamba     & 61.4 & 54.8 & 82.5 & 76.1 \\
PoolOnly       & 57.3 & 58.2 & 90.0 & 94.7 \\
LSTM           & 58.6 & 58.6 & 94.9 & 93.5 \\
GRU            & 60.2 & 58.7 & 94.8 & 93.8 \\
Transformer    & 63.8 & 60.5 & 95.5 & 95.2 \\
MeanPool       & 73.8 & 76.8 & 98.6 & 98.1 \\
\textbf{Mamba (ours)} & \textbf{80.6} & \textbf{84.1} & \textbf{99.2} & \textbf{99.5} \\
\bottomrule
\end{tabular}
\end{table}

The Mamba over segments wins every cell. Three gaps are diagnostic. The gap to MeanPool (segment-level Mamba retained, learned attention pool replaced by a plain mean) is $7$~pp on MuSiQue, isolating the contribution of the learned within-segment attention pool. The gap to TokenMamba (same Mamba module applied to concatenated tokens) is $19$ to $29$~pp on MuSiQue, isolating the contribution of placing the recurrence at the segment level rather than the token level. The gap to GRU and LSTM, which replace only the segment-level mixer with a two-layer classical recurrence of the same width, exceeds $20$~pp on MuSiQue in every cell. That gap isolates the selective state-space formulation from recurrence in general, and the LSTM carries slightly more trainable parameters than the Mamba mixer while still trailing it. On 2WikiMultihopQA, every architecture except LastTok and TokenMamba reaches at least $90\%$ R@$4$ and differences compress, but the Mamba still leads. Per-architecture analyses are in Appendix~\ref{app:router_details}.

\paragraph{Evidence relevance is recoverable from a model that cannot use it.}
Every architecture in Table~\ref{tab:router_arch} reads the frozen backbone with no adapter attached. Those same backbones answer MuSiQue at $14.4$ to $25.3$ F1 without adaptation (Table~\ref{tab:modulation_main}), yet a read-out of approximately $2.9$M parameters over their intermediate states ranks the supporting paragraphs at $80.6$ R@$4$ on Qwen and $99.2$ on 2WikiMultihopQA. The information required to identify the evidence is therefore already present in the representations of a model that does not act on it, and the role of the retrieval adapter is to decode that information rather than to learn it from scratch. A matched campaign supports this directly. Replacing the frozen base states with LoRA-adapted, TopLoRA-adapted or MaLoRA-adapted states changes supporting-paragraph recall by $0.7$ points on average, while the same adapters change end-task F1 on those cells by more than $36$ points (Appendix~\ref{app:base_vs_adapted}).

\paragraph{Comparison to classical retrieval baselines.}
BM25 \citep{Robertson2009ThePR} reaches R@$4{=}47.0$ on MuSiQue and $66.4$ on 2WikiMultihopQA. Qwen3-Embedding-8B \citep{zhang2025qwen3embeddingadvancingtext}, an $8$B-parameter dense retriever (larger than the backbone slice used here), reaches $75.5$ on MuSiQue via zero-shot cosine similarity. MaRA reaches $80.6$ and $99.2$ respectively, above both and above every retrieval baseline in Appendix~\ref{app:router_details} (Table~\ref{tab:retrieval_compare}). Both are external systems trained separately from the generator; MaRA instead reads the generator's own hidden states, so the advantage is not encoder size.

\begin{table*}[!t]
  \centering\small
  \setlength{\tabcolsep}{3pt}
  \caption{\textbf{Main results and component decomposition.} EM / F1 across two multi-hop reasoning datasets and three frozen backbones. Baselines use the full candidate set; retrieval-adapter rows use the top-$k$ retrieved paragraphs ($k{=}12$ MuSiQue, $k{=}4$ 2WikiMultihopQA). Best in each block in bold. The oracle row substitutes gold supporting paragraphs at the same $k$, as an upper bound.}
  \label{tab:main}
  \begin{tabular}{l ccc ccc}
    \toprule
    & \multicolumn{3}{c}{MuSiQue (EM / F1)} & \multicolumn{3}{c}{2WikiMultihopQA (EM / F1)} \\
    \cmidrule(lr){2-4}\cmidrule(lr){5-7}
    Method & Qwen & Llama & Gemma & Qwen & Llama & Gemma \\
    \midrule
    \multicolumn{7}{l}{\textit{No adapter}} \\
    Base, full context                & $5.1/14.4$ & $6.0/17.4$ & $13.8/25.3$ & $15.3/26.0$ & $20.3/31.9$ & $17.9/34.2$ \\
    \midrule
    \multicolumn{7}{l}{\textit{Static adapter baselines}} \\
    LoRA \citep{hu2022lora}           & $38.4/51.1$ & $41.8/54.4$ & $45.4/59.8$ & $71.7/77.3$ & $72.9/78.5$ & $71.5/77.2$ \\
    DoRA \citep{liu2024dora}          & $37.4/51.7$ & $41.7/56.0$ & $43.2/58.2$ & $70.7/77.3$ & $74.6/80.4$ & $70.7/76.9$ \\
    AdaLoRA \citep{zhang2023adaptive}  & $39.3/53.9$ & $44.4/59.7$ & $43.5/60.7$ & $68.0/73.7$ & $73.5/78.9$ & $75.3/80.4$ \\
    \midrule
    \multicolumn{7}{l}{\textit{Token-level modulated adapters}} \\
    TopLoRA \citep{li2025beyond}       & $41.5/55.6$ & $47.3/61.4$ & $43.8/59.4$ & $72.6/77.9$ & $75.1/80.4$ & $73.5/79.2$ \\
    MaLoRA                            & $\mathbf{42.6/56.4}$ & $\mathbf{48.1/62.3}$ & $\mathbf{48.7/63.7}$ & $\mathbf{73.7/79.3}$ & $\mathbf{78.1/83.3}$ & $\mathbf{77.1/82.2}$ \\
    \midrule
    \multicolumn{7}{l}{\textit{Modulator $+$ Retrieval Adapter}} \\
    LoRA $+$ BM25                       & $28.2/40.5$ & $29.6/41.1$ & $34.1/45.8$ & $51.3/56.3$ & $51.5/56.6$ & $50.0/55.4$ \\
    LoRA $+$ BGE-large (FT) \citep{10.1145/3626772.3657878} & $38.1/50.8$ & $40.6/54.0$ & $46.9/59.0$ & $75.2/80.8$ & $74.6/80.4$ & $66.2/73.2$ \\
    LoRA $+$ MaRA                       & $42.2/55.6$ & $43.4/55.8$ & $43.8/57.9$ & $75.4/81.0$ & $75.7/81.2$ & $74.6/80.3$ \\
    TopLoRA \citep{li2025beyond} $+$ MaRA & $44.2/58.4$ & $47.2/61.3$ & $42.6/57.7$ & $77.3/82.6$ & $77.8/83.1$ & $76.8/82.5$ \\
    \textbf{MaLoRA $+$ MaRA}          & $\mathbf{44.9/58.9}$ & $\mathbf{48.2/62.5}$ & $46.9/61.9$ & $\mathbf{77.9/83.3}$ & $\mathbf{79.9/85.1}$ & $\mathbf{80.1/85.2}$ \\
    \midrule
    \multicolumn{7}{l}{\textit{Modulator $+$ Oracle}} \\
    MaLoRA $+$ \emph{oracle gold evidence} & $46.7/61.2$ & $50.7/65.0$ & $51.8/66.8$ & $78.1/83.3$ & $80.8/85.9$ & $81.0/85.9$ \\
    \bottomrule
  \end{tabular}
\end{table*}

\section{Main Results}
\label{sec:results}

The full system combines the token-level modulator of Section~\ref{sec:modulation} (MaLoRA) with MaRA of Section~\ref{sec:routing}, both attached to a frozen low-rank adapter. MaRA restricts the language model's context to the top-$k$ retrieved segments before generation. Training and evaluation follow the setup of Sections~\ref{sec:modulation} and~\ref{sec:routing} on MuSiQue and 2WikiMultihopQA. All three backbones from Section~\ref{sec:modulation} (Qwen-2.5-7B, Llama-3.1-8B, Gemma-2-9B) remain frozen. All adapters share the canonical training pool described in Appendix~\ref{app:hparams}. Evaluation uses full-context decoding for the baselines and MaRA top-$k$ for the combined system. The number of retrieved segments $k$ is fixed once per dataset by supporting-paragraph recall on the validation split, not by end-task F1, and is then held fixed across all backbones and methods. Because a single $k$ is shared, it is set by the backbone whose router recalls least. For 2WikiMultihopQA ($10$ candidates) the default $k{=}4$ already attains $0.99$ recall or above on all three backbones. MuSiQue has $20$ candidates and harder multi-hop retrieval, and Gemma's router is the binding constraint: $k{=}12$ is required to bring it to $0.94$. Per-backbone recall at each $k$ is in Appendix~\ref{app:per-cell}. End-task scores are exact match (EM) and F1.

Table~\ref{tab:main} reports the full system against the LoRA baseline across the grid. The MaLoRA$+$MaRA system improves over LoRA on every cell, by $+2.1$ to $+8.1$ F1 ($+6.4$ F1 on average, $+10.0\%$ relative): the largest margin is on MuSiQue Llama ($54.4 \to 62.5$, $+8.1$) and the smallest on MuSiQue Gemma ($59.8 \to 61.9$, $+2.1$); on 2WikiMultihopQA the gains are $+6.0$/$+6.6$/$+8.0$ F1 across Qwen/Llama/Gemma. The mechanism analyses for the two components are in Section~\ref{sec:modulation} (modulator alone, Table~\ref{tab:modulation_main}) and Section~\ref{sec:routing-ablation} (MaRA architecture, Table~\ref{tab:router_arch}).

\begin{figure*}[t]
  \centering
  \includegraphics[width=0.95\linewidth]{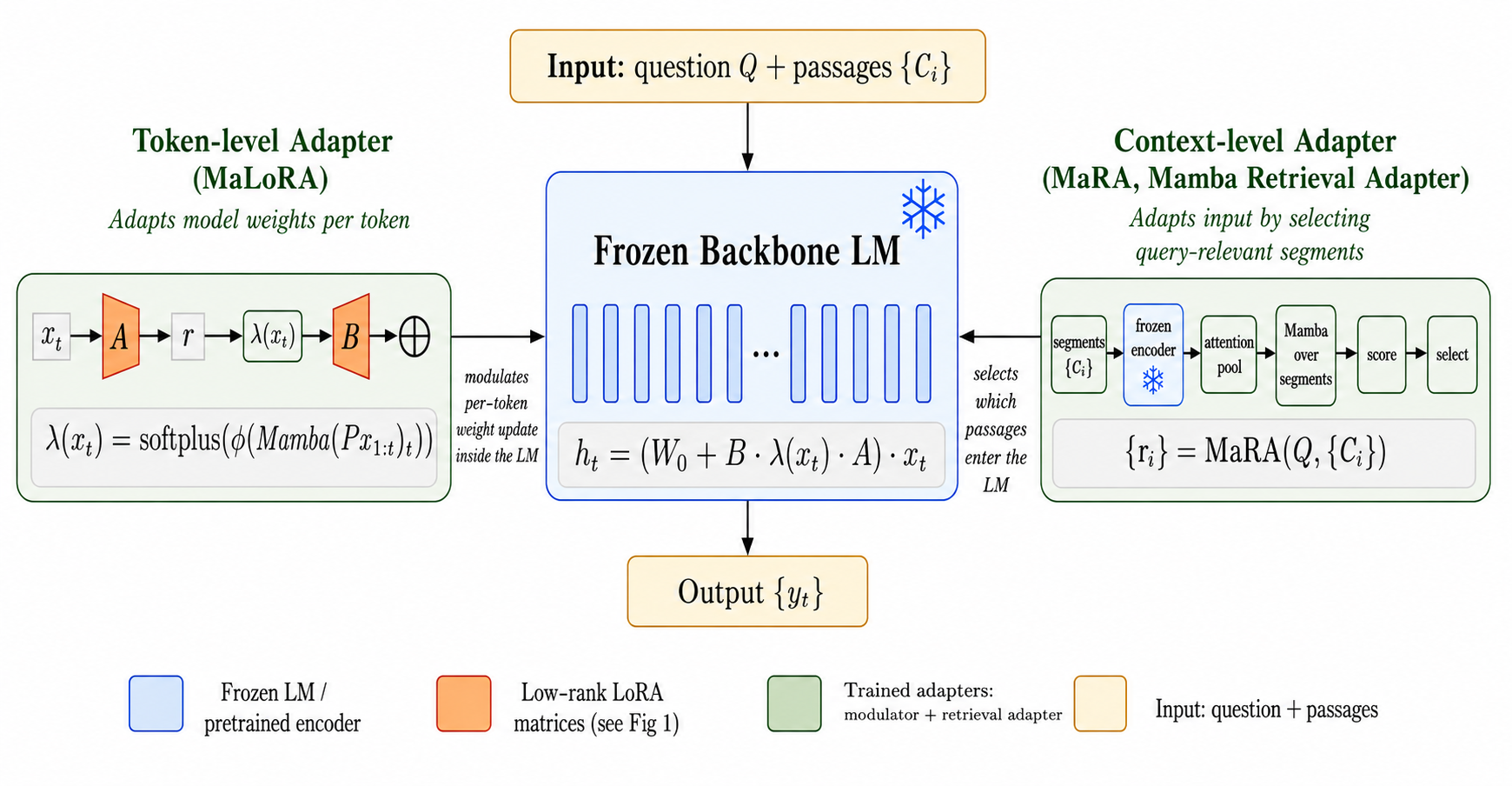}
  \caption{\textbf{Token-level and context-level adaptation around a frozen LM.} The token-level adapter (MaLoRA, left) modulates the LoRA scaling factor per token, inside the language model. The context-level adapter (MaRA, right) selects which query-relevant segments enter the language model, before generation. Same frozen backbone, different granularities, different targets.}
  \label{fig:overview}
\end{figure*}

\paragraph{Two-level selective state-space adaptation around a frozen LM.}
Figure~\ref{fig:overview} places both adapters around the shared frozen backbone. They differ in what they read, what they emit, and when they act. MaLoRA reads the token hidden state $x_t$ inside every adapted projection and emits a scalar that rescales the low-rank update during generation. MaRA reads per-segment summaries from the first $K$ layers of the same backbone and emits a selection over candidate paragraphs before generation begins. Neither can express the other: a per-token scalar cannot remove a paragraph from the context, and a per-segment selection cannot vary adapter strength across the tokens it admits.

The measured gains behave accordingly. Token-level modulation lifts LoRA in every cell of Table~\ref{tab:main}, by $+2.0$ to $+7.9$ F1. Retrieval alone lifts it by $-1.9$ to $+4.5$ F1. Summed per cell, the two account for the full-system gain with a mean overlap of $6\%$, so the two paths are close to additive.

The variability is one-sided. Modulation is positive on all six cells. Retrieval is the component whose value depends on the task, tracking how hard the selection problem is and how well the frozen backbone's states separate evidence from distractors. On 2WikiMultihopQA ($10$ candidates) MaRA reaches the gold-evidence ceiling on all three backbones, coming within $0.0$, $0.8$ and $0.7$ F1 of MaLoRA$+$oracle. On MuSiQue ($20$ candidates, chains up to length four) $+2.3$ to $+4.9$ F1 of headroom remains, and on Gemma, whose router recalls least (Table~\ref{tab:recall_per_bb}), retrieval alone costs $1.9$ F1. Modulation is unaffected by that difficulty because it never sees the candidate set. The remaining MuSiQue headroom is one of precision rather than recall. With the gold supporting paragraphs alone and no padding to $k$, MaLoRA reaches $68.8$, $71.1$ and $74.1$ F1 on Qwen, Llama and Gemma. Those figures exceed the oracle row of Table~\ref{tab:main} by $7.6$, $6.1$ and $7.3$ F1, since the oracle pads to $k$ with distractors. Reducing $k$ to $8$ does not recover the gap. It loses $1.0$ F1 on average, because recall falls faster than precision improves (Appendix~\ref{app:per-cell}).

\paragraph{Efficiency.}
Parameter overhead, training time, peak memory, and inference latency are detailed in Appendix~\ref{app:param_efficiency} (Table~\ref{tab:efficiency}).


\section{Conclusion}
\label{sec:conclusion}

A family of selective state-space adapters was introduced for frozen language models, operating at the token level (MaLoRA) and at the context-segment level (MaRA). Across three backbones and two multi-hop reasoning datasets, the combined system improved over LoRA on every cell of the $3{\times}2$ grid, by $+6.4$ F1 on average ($+10.0\%$ relative). The two components read different inputs and emit different outputs at different points in the forward pass, so neither can express the other and their gains largely accumulate. Their relative value varies with the task: on 2WikiMultihopQA the combined system reaches the gold-evidence ceiling, whereas on MuSiQue a gap of $+2.3$ to $+4.9$ F1 to that ceiling remains.

Two directions extend this work. \textbf{(i)}~Transfer to reasoning beyond multi-hop QA requires a task-appropriate notion of segment. Token-level modulation already shows gains on commonsense and mathematical reasoning (Appendices~\ref{app:commonsense} and~\ref{app:additional}), whereas a solution-step segmentation for MaRA did not yield a clean lift. Defining the segment is therefore the open problem. \textbf{(ii)}~MaRA was used here only for segment selection. A natural extension injects its per-segment state directly into the generator through steering vectors or activation patching at chosen layers.

\newpage

\section*{Limitations}
\label{sec:limitations}

Evaluation of the full MaLoRA$+$MaRA system is restricted to supervised multi-hop question answering. Token-level modulation alone is additionally evaluated on long-context (Table~\ref{tab:modulation_main}), commonsense (Appendix~\ref{app:commonsense}) and mathematical reasoning (Appendix~\ref{app:additional}). Segment-level retrieval did not extend to mathematics. A solution-step variant of the retrieval adapter was attempted and did not yield a clean lift, which is reported as future work rather than a generalisation claim. Other reasoning settings, including code generation, few-shot in-context learning without training, and conversational agents with evolving state, remain out of scope of the present empirical evaluation.

The main grid spans the $7$--$9$B parameter range. The token-level comparison is additionally run on Qwen-2.5-3B at a single seed (Appendix~\ref{app:small_backbone}), where the ordering is preserved, but no backbone below $3$B or above $9$B is tested. MaRA is additionally trained at $3$B, where recall on 2WikiMultihopQA matches the $7$B backbone ($99.1$ against $99.2$ R@$4$) but MuSiQue recall falls from $80.6$ to $72.0$. Retrieval quality therefore tracks backbone capacity on the harder selection problem, as expected of a component that reads the backbone's own representations. The full system is not evaluated end-to-end at that scale, so how the two contributions balance at smaller capacity is only partially measured.

The retrieval adapter is trained with supervised binary labels over the supporting paragraphs for each example. Settings without such labels, or with noisy labels, are out of scope. Distantly supervised or self-supervised variants of the retrieval objective would be needed to extend the method to such settings.

The training schedule of both the adapter and the retrieval adapter is sensitive to the choice of backbone. Gemma-2-9B required a smaller learning rate and a tighter gradient-norm clip than Qwen-2.5-7B and Llama-3.1-8B to train stably. Practitioners applying this method to a new backbone should expect to retune these two values. Each main-result cell consumed approximately $10$ to $20$ GPU-hours of training on a single A100-class device (Table~\ref{tab:efficiency}). With five adapter methods across the $3 \times 2$ grid and up to three seeds per cell, the full reported set required on the order of $1{,}000$ GPU-hours, a non-trivial barrier to seed-replication at larger scales.

The retrieval adapter does not help uniformly. On MuSiQue with Gemma-2-9B it costs $1.9$ F1 relative to the same adapter without retrieval, the only such cell in the grid. A single $k$ is held fixed across backbones for comparability, which penalises the backbone whose router recalls least (Appendix~\ref{app:per-cell}); a per-backbone $k$ would recover part of that loss but would confound the comparison.

Generation is performed with key-value caching enabled. The token-level modulator carries no incremental-decode state across cached steps, so its recurrence is restarted at each generated position. The prompt is processed in a single forward pass and is therefore modulated over its full trajectory; only the generated answer tokens are affected. A controlled comparison with caching disabled was run on all six cells at seed $42$. F1 moved by $-0.5$ to $+1.1$, a mean of $+0.2$, with no consistent direction. An incremental-decode path for the modulator is left to future work.

The selective state-space blocks use \texttt{mambapy}~$1.2.0$, a pure-PyTorch implementation of the Mamba-1 (S6) recurrence, rather than the fused CUDA kernels of the reference implementation. The wall-clock figures in Appendix~\ref{app:param_efficiency} therefore understate what an optimised kernel would achieve, and no efficiency claim is made on their basis.

Training is staged rather than end-to-end. The modulated adapter (LoRA $+$ MaLoRA) is trained first on the target task, and MaRA is then trained against the unadapted backbone, with no modulator attached. Joint end-to-end optimisation of the two stages is left as future work. Modest additional gains are expected from co-adapting MaRA's segment selection to the modulated hidden states, but staged training suits the model-attached design: MaRA reads frozen hidden states, so its training cost stays low.

\section*{Ethics Statement}
\label{sec:ethics}

All datasets used in this work are publicly released benchmarks for multi-hop question answering. No human-subject data is collected or annotated. The backbones used (Qwen-2.5-7B \citep{qwen2025qwen25technicalreport}, Llama-3.1-8B \citep{grattafiori2024llama3}, Gemma-2-9B \citep{riviere2024gemma2}) are released under their respective licenses and used in accordance with them. No known dual-use concerns arise from the contributions of this paper: the modulator and retrieval adapter are general-purpose adaptation components and do not enable new harmful capabilities beyond those already present in the underlying language models. The compute footprint is documented in the Limitations section above.

\paragraph{Use of AI assistants.}
AI-based writing assistants (large language model chat assistants) were used for prose polishing, surfacing inconsistencies between sections, drafting LaTeX scaffolding for tables, and code-assisted error checking during the experimental pipeline (e.g., auditing eval configurations for mismatched checkpoints). All research design, methodological decisions, experiments, analyses, and final wording remain the responsibility of the human authors.

\bibliography{references}

\appendix

\section{Hyperparameter and training details}
\label{app:hparams}

\paragraph{Datasets.}
MuSiQue \citep{trivedi-etal-2022-musique} (denoted MQ) has $2$-, $3$-, and $4$-hop compositional questions over $20$ candidate paragraphs. 2WikiMultihopQA \citep{ho-etal-2020-constructing} (2Wiki) has compositional, comparison, inference, and bridge-comparison questions over $10$ candidate paragraphs. The fraction of paragraphs that are supporting is small in both.

\paragraph{Canonical training pool.}
A single shared training pool per dataset is constructed for fair cross-backbone comparison, capped at $10{,}000$ examples and yielding $9{,}998$ after filtering. An example is admitted if the prompt length under both the Qwen and Llama tokenisers is at most the dataset's sequence-length budget ($2048$ for 2Wiki, $4096$ for MQ), and the example must satisfy a chunk-detection check under both tokenisers. Remaining examples are sampled with stratification by hop count (MQ) or by question type (2Wiki) under a fixed random seed of $42$. Every model is trained on the same pool, so cross-backbone differences are not attributable to different training subsets.

\paragraph{Evaluation splits.}
End-task evaluation uses the released validation splits. 2WikiMultihopQA takes the first $3{,}000$ examples (\texttt{dataset[:3000]}). MuSiQue uses the full answerable subset of its validation split ($2{,}417$ examples after filtering to \texttt{answerable=True}); see \texttt{eval\_musique.py} for the procedure. A negligibly small number of prompt cases that exceed the model's sequence budget are skipped rather than truncated. Truncation would silently drop candidate paragraphs, including supporting ones, and would confound retrieval quality with context loss. The number skipped differs slightly across backbones because tokenisation does. Within a cell every method is evaluated on the same examples. All evaluations use beam search of width $4$.

\paragraph{Adapter training (LoRA, modulator).}
LoRA rank $r{=}16$; the standard LoRA scaling is absorbed into the initialization of $A$ and $B$ (Eq.~\ref{eq:lora}). Adapters are attached to $\{q\_\mathrm{proj}, k\_\mathrm{proj}, v\_\mathrm{proj}, \mathrm{up}\_\mathrm{proj}, \mathrm{down}\_\mathrm{proj}\}$. Training is $2$ epochs with learning rate $2{\times}10^{-4}$, weight decay $0$, $100$ warmup steps, bf16 mixed precision. The per-device batch size and the gradient-accumulation factor vary by cell to fit memory. On MuSiQue the effective batch size is $4$ throughout, applied as $2{\times}2$ on Qwen-2.5-7B and Llama-3.1-8B and as $1{\times}4$ on Gemma-2-9B. On 2WikiMultihopQA it is $8$ on Qwen-2.5-7B and Llama-3.1-8B, applied as $4{\times}2$. On Gemma-2-9B with 2WikiMultihopQA the LoRA, AdaLoRA, DoRA and MaLoRA runs use $1{\times}4$ and the TopLoRA run uses $1{\times}8$, which is the one cell in which the compared methods do not share an effective batch size. The LoRA and MaLoRA settings are matched in every cell. The modulated variants (Section~\ref{sec:modulation}) use the same schedule, loss, and module set as LoRA. The MaLoRA modulator uses Mamba state dimension $d_s{=}16$, distinct from the $d_s{=}64$ used by MaRA below. Modulator-only parameters add approximately $54\%$ (TopLoRA) or approximately $56\%$ (MaLoRA) over the LoRA parameter count; see Appendix~\ref{app:param_efficiency}.

\paragraph{Adapter training on Gemma.}
Gemma-2-9B uses a safer schedule: learning rate $1{\times}10^{-4}$, $\max\|g\|_2 = 0.5$, $300$ warmup steps. All other settings match the table above. This applies uniformly to LoRA, DoRA, AdaLoRA, and the modulated variants.

\paragraph{Retrieval adapter training.}
MaRA (Section~\ref{sec:routing}) is trained for $2$ epochs with batch size $2$ and gradient-accumulation $2$. Qwen and Llama backbones use learning rate $5{\times}10^{-4}$ and $\max\|g\|_2 = 1.0$. The chunk encoder uses the first $K$ transformer layers of the backbone with frozen weights, with $K{=}16$ for Qwen, $K{=}12$ for Llama, and $K{=}20$ for Gemma on MuSiQue and $K{=}16$ on 2WikiMultihopQA. The recurrence width is $D{=}256$, attention-pool dimension is $256$, scoring-head MLP hidden dimension is $256$, Mamba state dim $d_s{=}64$. Positive-class BCE weight is $8.0$; pairwise margin loss weight $0.5$ with margin $1.0$. MaRA has $2.86$M trainable parameters on Qwen and Gemma and $3.13$M on Llama.

\paragraph{Retrieval adapter training on Gemma.}
MaRA on Gemma uses the safer schedule: learning rate $1{\times}10^{-4}$, $\max\|g\|_2 = 0.5$. Other settings unchanged.

\paragraph{Seeds and checkpoint selection.}
The LoRA baseline, both token modulators (TopLoRA, MaLoRA) and every modulator$+$MaRA row are trained with three random seeds ($42$, $43$, $44$), and the corresponding entries in Table~\ref{tab:main} and Table~\ref{tab:modulation_main} are means over those three seeds with no best-seed selection. DoRA is run at three seeds on MuSiQue with Llama-3.1-8B and Gemma-2-9B and at a single seed on the remaining four cells; AdaLoRA, the BM25 and BGE-large retrieval rows, and the oracle row are single-seed. The gold-only evaluation reported in Section~\ref{sec:results} uses three seeds on Qwen-2.5-7B and Llama-3.1-8B and one seed on Gemma-2-9B. For analyses that operate on a single checkpoint (the state-usage probe of Appendix~\ref{app:state_probe}, the per-projection / per-layer marginals of Appendix~\ref{app:proj_layer_marginals}, and the per-token region analysis of Appendix~\ref{app:region_lambda}), the seed with the highest validation F1 is used per cell; this is a methodological choice (these analyses require a specific checkpoint) and does not affect any reported headline metric.

\section{Efficiency breakdown: parameters, training time, peak memory, inference latency}
\label{app:param_efficiency}

This appendix combines the four cost dimensions of each adapter into a single view: \textbf{(i)} trainable parameter count, \textbf{(ii)} end-to-end training time, \textbf{(iii)} peak GPU memory during training, and \textbf{(iv)} inference latency. Table~\ref{tab:efficiency} reports each dimension averaged over the six cells of Table~\ref{tab:main} ($3$ backbones $\times$ $2$ datasets); per-cell numbers are recorded in the per-run \texttt{metrics.json} files released with the code, and the training configuration for every cell is committed under \texttt{configs/canonical\_pool/}.

\paragraph{Adapter training (modulators against static baselines).}
\textbf{DoRA is the most expensive among the static baselines:} its weight-decomposed reparameterisation takes $1.90\times$ the LoRA training time on average ($20.3$ h vs.\ $10.8$ h), and $30.4$ h on MuSiQue Gemma. AdaLoRA is closer to LoRA at $1.04\times$. \textbf{Both token modulators are cheaper than DoRA at training:} TopLoRA trains at $0.95\times$ LoRA time, and MaLoRA at $1.16\times$ LoRA time, the cost of adding a recurrent module. Peak GPU memory is within approximately $5\%$ of LoRA for every method: peak training memory is dominated by the frozen backbone and activations, not by the adapter delta at rank $r{=}16$.

\paragraph{Retrieval adapter (MaRA) training.}
MaRA is trained as a separate stage with the language model frozen and adds only the segment-level mixer and scoring heads (approximately $2.9$ to $3.1$M parameters, $\approx 10\%$ of the LoRA adapter). Average training time is $\approx 2.5$ h per (backbone, dataset) cell, peak GPU memory $\approx 12$ GB. Because MaRA reuses the first $K$ frozen layers of the backbone as encoder and trains only the small mixer on top, both its training time and its memory footprint are well below those of the LoRA-side adapter.

\paragraph{Inference latency.}
Latency is measured on a dedicated idle GPU with every configuration run sequentially, so no measurement competes with another job; $n{=}200$ validation examples per configuration at beam $4$ and rank $r{=}16$. Per-sample cost is flat from $n{=}50$ onward, so the means in Table~\ref{tab:latency} are not warm-up artefacts. The two components are separated by running each cell twice, differing only in whether the retrieval adapter runs: once selecting $k$ paragraphs at random (generation only), and once with the full chained adapter.

\begin{table}[!h]
  \centering\small
  \setlength{\tabcolsep}{4pt}
  \caption{\textbf{Inference latency} (s/sample; $n{=}200$, beam $4$, dedicated idle GPU, sequential runs). \emph{Gen.} retains $k$ paragraphs without running the retrieval adapter; \emph{$+$MaRA} adds the $N_p{=}4$ chained pre-pass; \emph{Pre-pass} is their difference.}
  \label{tab:latency}
  \begin{tabular}{ll ccc}
    \toprule
    DS & BB & Gen. & $+$MaRA & Pre-pass \\
    \midrule
    MQ & Qwen  & $1.88$ & $5.97$ & $4.09$ \\
    MQ & Llama & $2.23$ & $5.21$ & $2.98$ \\
    MQ & Gemma & $3.51$ & $9.12$ & $5.61$ \\
    \midrule
    2Wiki & Qwen  & $1.27$ & $3.92$ & $2.65$ \\
    2Wiki & Llama & $1.31$ & $3.41$ & $2.10$ \\
    2Wiki & Gemma & $1.86$ & $4.67$ & $2.80$ \\
    \bottomrule
  \end{tabular}
\end{table}

\emph{(a) LM forward.} Generation cost is set by the backbone and the retained context length, not by the adapter choice: at fixed backbone the spread across adapters is below $5\%$, since the modulator adds one scalar matmul per token plus a $d_s{=}16$-state Mamba step, negligible against the backbone's own matmuls. Gemma-2-9B is the slowest backbone, in part because of logit soft-capping.

\emph{(b) MaRA segment scoring (pre-pass).} The retrieval adapter re-encodes every candidate through the first $K$ frozen layers once per chained pass, and each pass conditions on a longer anchor prefix than the one before it, so later passes cost more than earlier ones. At $N_p{=}4$ the pre-pass adds $2.1$ to $5.6$ s/sample, which is $1.3$ to $2.2$ times the generation it precedes. This is the dominant inference cost of the system and the main practical price of context-level adaptation. Fewer passes would reduce it close to proportionally; $N_p$ is set to match the pass budget the adapter was trained with rather than tuned for latency.

\begin{table*}[!h]
  \centering\small
  \setlength{\tabcolsep}{4pt}
  \caption{\textbf{Efficiency breakdown.} Trainable parameters, training time (mean over $3$ backbones $\times$ $2$ datasets), and peak GPU memory during training, each averaged over the same six cells. ``$\times$LoRA'' is the ratio to the LoRA baseline's training time on the same six cells; ratios are computed from unrounded training times. DoRA's $1.90\times$ training time is the most expensive baseline; MaLoRA costs $1.16\times$, the overhead of the recurrent state-space module. MaRA is trained separately with the language model frozen and adds $2.9$ to $3.1$M parameters and $\approx 2.5$ h training per cell.}
  \label{tab:efficiency}
  \begin{tabular}{lccccc}
    \toprule
    Method & Params (M) & $\Delta$LoRA & Train (h) & Train $\times$LoRA & Peak GPU (GB) \\
    \midrule
    \multicolumn{6}{l}{\textit{Static adapter baselines}} \\
    LoRA \citep{hu2022lora}            & $30.7$ & --       & $10.8$ & $1.00\times$ & $33.8$ \\
    AdaLoRA \citep{zhang2023adaptive}   & $46.1$ & $+50\%$  & $11.1$ & $1.04\times$ & $34.0$ \\
    DoRA \citep{liu2024dora}           & $31.6$ & $+3\%$   & $\mathbf{20.3}$ & $\mathbf{1.90\times}$ & $33.8$ \\
    \midrule
    \multicolumn{6}{l}{\textit{Token modulators}} \\
    $+$ TopLoRA \citep{li2025beyond}     & $47.4$ & $+54\%$ & $10.3$ & $0.95\times$ & $35.2$ \\
    $+$ MaLoRA                          & $47.9$ & $+56\%$ & $12.4$ & $1.16\times$ & $35.3$ \\
    \midrule
    \multicolumn{6}{l}{\textit{Retrieval adapter (separate stage, language model frozen)}} \\
    $+$ MaRA (alone)                    & $3.1$  & $+10\%$ & $2.5$  & $0.23\times$ & $12.0$ \\
    \bottomrule
  \end{tabular}
\end{table*}

The combined picture: among static baselines, MaLoRA trains at $1.16\times$ the LoRA baseline, slightly above AdaLoRA ($1.04\times$), while delivering a larger F1 gain (Table~\ref{tab:main}), and is $1.63\times$ faster to train than DoRA. Adding MaRA on top costs a separate $2.5$ h per cell with a small ($12$ GB) memory footprint, well below the main adapter's profile. At inference, the choice among token modulators is essentially free at rank $16$ and beam $4$; the retrieval pre-pass is not, and dominates inference cost (Table~\ref{tab:latency}).

\section{Stateless modulator head: activation choice}
\label{app:ablations}

This appendix isolates the per-token \emph{stateless} modulator head — the scalar $\lambda(x_t)=f(\phi(P x_t))$ family without the Mamba recurrence — under three activation choices (softplus, sigmoid, tanh). Table~\ref{tab:scalar_diag} reports the matrix under the Table~\ref{tab:modulation_main} protocol, full-context, on MuSiQue and 2WikiMultihopQA across all three backbones. This ablation does \emph{not} include the Mamba recurrence; MaLoRA in the main paper is the softplus-scalar variant with a Mamba block added on top (Section~\ref{sec:modulation-method}). Its Table~\ref{tab:modulation_main} numbers are higher than the corresponding stateless row on five of six cells; on Llama 2WikiMultihopQA the stateless variant is marginally higher ($84.1$ vs.\ $83.3$), within the per-seed standard deviation ($\sigma \le 0.9$ for Llama 2WikiMultihopQA in Table~\ref{tab:seed_std}).

The three activations land within approximately $2$ F1 of each other on every cell, with no consistent winner across backbones. Softplus is preferred over sigmoid because its unbounded positive range admits both suppression and amplification, the latter empirically observed for MaLoRA on Gemma-2-9B (Table~\ref{tab:region_lambda}, marker/question tokens with $\lambda > 1$); a sigmoid-bounded modulator $(\lambda \in (0,1))$ cannot represent this amplification, and its near-equal end-task F1 reflects that on these benchmarks the modulator is suppression-dominated while the capacity to amplify is exercised mainly by the Gemma cell.

\begin{table}[!h]
  \centering\small
  \setlength{\tabcolsep}{4pt}
  \caption{\textbf{Stateless modulator-head activation ablation} (F1, full context, full validation split, beam $4$; scalar output form). The per-token head is a linear $\phi$ over $P x_t$ with no recurrence. $^{\dagger}$Tanh on MuSiQue Gemma-2-9B fails to learn (F1 $26.3$ vs.\ $\ge 63.2$ for the other activations); the same tanh head reaches healthy F1 on 2WikiMultihopQA Gemma ($83.1$) and on Qwen/Llama.}
  \label{tab:scalar_diag}
  \begin{tabular}{l ccc ccc}
    \toprule
    & \multicolumn{3}{c}{MuSiQue F1} & \multicolumn{3}{c}{2Wiki F1} \\
    \cmidrule(lr){2-4}\cmidrule(lr){5-7}
    Activation & Qw & Ll & Ge & Qw & Ll & Ge \\
    \midrule
    softplus & $55.1$ & $61.7$ & $63.2$ & $78.5$ & $84.1$ & $81.8$ \\
    sigmoid  & $54.2$ & $63.0$ & $64.5$ & $79.0$ & $83.9$ & $83.7$ \\
    tanh     & $55.2$ & $61.8$ & $\mathbf{26.3}^{\dagger}$ & $78.4$ & $83.0$ & $83.1$ \\
    \bottomrule
  \end{tabular}
\end{table}

\paragraph{Stateful diagonal output form.}
Table~\ref{tab:scalar_diag} holds the output form scalar. Table~\ref{tab:stateful_diag} 
varies the output form with the Mamba recurrence in place, comparing the stateful-diagonal 
variant against MaLoRA over the same three seeds. The scalar form is higher on three of the 
four cells. In the single cell where the diagonal form leads, MuSiQue with Llama-3.1-8B, it 
carries twice the seed variance ($\sigma$ of $3.6$ against $1.8$). On 2WikiMultihopQA with 
Qwen-2.5-7B the diagonal form is both lower and an order of magnitude less stable ($\sigma$ 
of $5.8$ against $0.5$). The scalar output form is therefore retained throughout.

\begin{table}[!h]
  \centering\small
  \setlength{\tabcolsep}{1.2pt}
  \caption{\textbf{Stateful diagonal against scalar output} (F1, mean $\pm$ standard deviation over seeds $42$, $43$, $44$; full context, beam $4$). Both variants use the Mamba recurrence and differ only in the output form of $\lambda$.}
  \label{tab:stateful_diag}
  \begin{tabular}{l cc cc}
    \toprule
    & \multicolumn{2}{c}{MuSiQue} & \multicolumn{2}{c}{2Wiki} \\
    \cmidrule(lr){2-3}\cmidrule(lr){4-5}
    Output form & Qw & Ll & Qw & Ll \\
    \midrule
    Diagonal        & $55.6${\scriptsize$\pm1.4$} & $\mathbf{63.2}${\scriptsize$\pm3.6$} & $74.4${\scriptsize$\pm5.8$} & $82.1${\scriptsize$\pm0.3$} \\
    Scalar(MaLoRA) & $\mathbf{56.4}${\scriptsize$\pm1.1$} & $62.3${\scriptsize$\pm1.8$} & $\mathbf{79.3}${\scriptsize$\pm0.5$} & $\mathbf{83.3}${\scriptsize$\pm0.7$} \\
    \bottomrule
  \end{tabular}
\end{table}

\paragraph{Per-seed variance of the three token modulators.}
The LoRA, TopLoRA, MaLoRA and modulator-plus-MaRA entries in Table~\ref{tab:main} are means over three random seeds ($42$, $43$, $44$); DoRA and AdaLoRA follow the seed protocol given in Appendix~\ref{app:hparams}. Table~\ref{tab:seed_std} reports the corresponding per-seed standard deviations for the token modulators. Per-cell variance is small across all three token modulators ($\sigma \le 2.6$ F1 in every cell).

\begin{table}[!h]
  \centering\footnotesize
  \setlength{\tabcolsep}{2.5pt}
  \caption{\textbf{Per-seed F1 standard deviation} (over $3$ seeds: $42$, $43$, $44$).}
  \label{tab:seed_std}
  \begin{tabular}{l cccccc}
    \toprule
    & \multicolumn{3}{c}{MuSiQue} & \multicolumn{3}{c}{2Wiki} \\
    \cmidrule(lr){2-4}\cmidrule(lr){5-7}
    Method & Qw & Ll & Ge & Qw & Ll & Ge \\
    \midrule
    TopLoRA \citep{li2025beyond} & $2.5$ & $2.4$ & $2.6$ & $0.8$ & $0.9$ & $2.2$ \\
    MaLoRA                      & $1.1$ & $1.8$ & $2.3$ & $0.5$ & $0.7$ & $2.1$ \\
    \bottomrule
  \end{tabular}
\end{table}

\section{Per-projection and per-layer modulation marginals}
\label{app:proj_layer_marginals}
\begin{figure}[!h]
  \centering
  \includegraphics[width=0.99\linewidth]{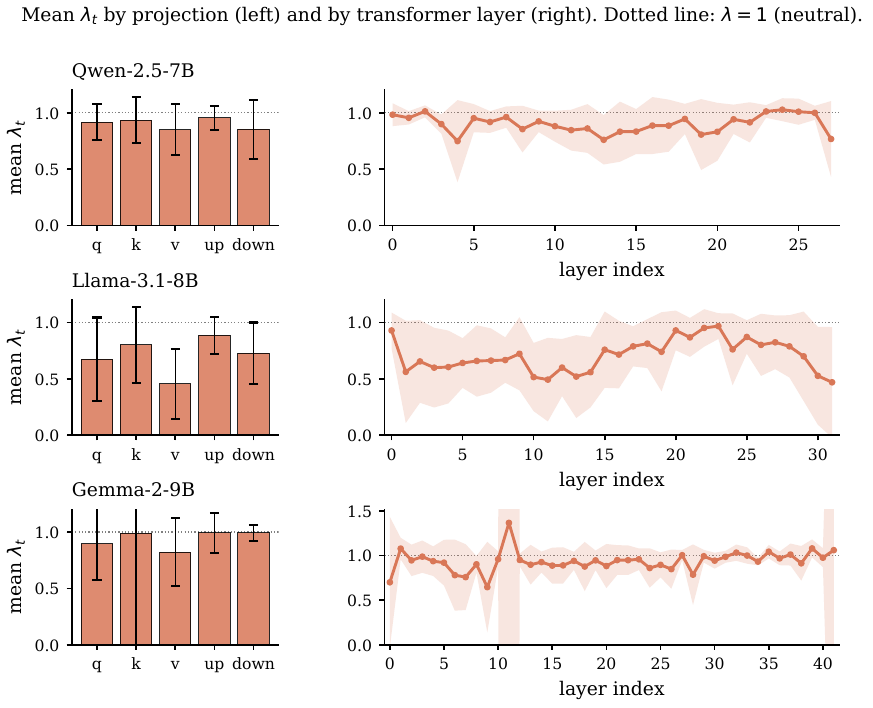}
  \caption{\textbf{Where the modulator suppresses.} Mean $\lambda(x_t)$ by projection type (left) and transformer layer (right), aggregated over the full MuSiQue validation set. Three cells: MaLoRA on Qwen-2.5-7B, Llama-3.1-8B, and Gemma-2-9B.}
  \label{fig:layer_proj}
\end{figure}

Figure~\ref{fig:layer_proj} reports the mean of $\lambda(x_t)$ broken down by projection type and by transformer layer for the three MaLoRA-backbone cells: Qwen-2.5-7B, Llama-3.1-8B, and Gemma-2-9B. The figure is referenced from Section~\ref{sec:modulation-behavior}. Suppression is deepest in $v\_\mathrm{proj}$ in all three cells. The final transformer layer is suppressed on Qwen-2.5-7B and Llama-3.1-8B, but not on Gemma-2-9B, whose per-layer means stay close to one throughout.

\section{Modulation by token role: extended discussion}
\label{app:region_lambda}

The per-region means of $\lambda(x_t)$ reported in Table~\ref{tab:region_lambda} (main body) are aggregated over the full MuSiQue validation set, across the three MaLoRA-backbone cells. Tokens are tagged as \emph{marker} (format/separator tokens), \emph{question} (the query), \emph{para\_support} / \emph{para\_distractor} (supporting vs.\ non-supporting paragraph content), and \emph{context-other}.

Two findings are robust. First, \textbf{$\lambda$ on supporting and distractor paragraphs differs by at most $0.002$ in every cell}: the modulator does not separate relevant from irrelevant evidence. This is expected, since identifying supporting evidence requires the multi-hop reasoning that the modulator precedes; evidence selection is instead handled by MaRA (Section~\ref{sec:routing}). Second, the modulator instead allocates modulation by \textbf{token role}: on Qwen and Llama, structural (marker) and query tokens are suppressed more than paragraph content, consistent with the per-token view in Figure~\ref{fig:token_heatmap}. The exception is MaLoRA on Gemma-2-9B, which \emph{amplifies} query and marker tokens ($\lambda \approx 1.2$--$1.7$) rather than suppressing them, a backbone-specific behavior. The shared pattern (relevance-blind, role-sensitive) explains why the modulator and MaRA are non-redundant: the modulator reshapes the low-rank update by token role and cannot perform evidence selection, so MaRA contributes an orthogonal capability.

\section{State-usage probe: per-cell results}
\label{app:state_probe}

This appendix expands the state-usage check from Section~\ref{sec:modulation-behavior}. For each Mamba block in a trained MaLoRA cell, the block's full stateful output $z_t = \mathrm{Mamba}(P x_{1:t})_t$ is compared against a stateless variant $z_t^{\mathrm{stateless}}$ in which each position is processed independently with the recurrent state zeroed. The relative discrepancy is
\begin{equation*}
\rho_t \;=\; \frac{\bigl\|z_t - z_t^{\mathrm{stateless}}\bigr\|_2}{\|z_t\|_2}.
\end{equation*}
A value of $\rho_t \approx 0$ at every position would indicate that the recurrent state contributes nothing and that the Mamba block could be replaced by a stateless head; $\rho_t > 0$ means the state genuinely accumulates information up to position $t$. The probe is run on $5$ validation samples and $10$ randomly sampled MambaBlocks per cell, on the six canonical MaLoRA cells, using the locked best-seed checkpoints from the manifest. Per-cell mean $\rho_t$ at sampled positions $t \in \{1, 10, 10^2, 10^3\}$ is in Table~\ref{tab:state_probe}. The aggregate across the six cells rises from $0.13$ at position $1$ to $0.61$--$0.63$ across the rest of the prompt, with every cell above $0.45$ from $t{=}10$ onward; the recurrence carries useful trajectory-level information rather than acting decoratively.

\begin{table}[!h]
  \centering\small
  \setlength{\tabcolsep}{5pt}
  \caption{\textbf{State-usage probe: mean $\rho_t$ by token position.} Averaged over $10$ sampled Mamba blocks and $5$ validation prompts; larger $\rho_t$ = more state contribution.}
  \label{tab:state_probe}
  \begin{tabular}{ll cccc}
    \toprule
    DS & BB & $t{=}1$ & $t{=}10$ & $t{=}10^2$ & $t{=}10^3$ \\
    \midrule
    MQ & Qwen  & $0.18$ & $0.55$ & $0.56$ & $0.62$ \\
    MQ & Llama & $0.10$ & $0.60$ & $0.61$ & $0.65$ \\
    MQ & Gemma & $0.09$ & $0.76$ & $0.69$ & $0.61$ \\
    \midrule
    2Wiki & Qwen  & $0.23$ & $0.59$ & $0.64$ & $0.56$ \\
    2Wiki & Llama & $0.05$ & $0.47$ & $0.53$ & $0.52$ \\
    2Wiki & Gemma & $0.14$ & $0.71$ & $0.78$ & $0.78$ \\
    \midrule
    \multicolumn{2}{l}{\emph{Mean (6 cells)}} & $0.13$ & $0.61$ & $0.63$ & $0.62$ \\
    \bottomrule
  \end{tabular}
\end{table}

\section{Recurrent family at both levels}
\label{app:recurrent}

The selective state-space module is replaced by a GRU and by an LSTM, at the token level (the MaLoRA modulator of Section~\ref{sec:modulation-method}) and at the segment level (the MaRA mixer of Section~\ref{sec:routing}). Everything else is held fixed. At the token level the projection $P$, the scalar output, the softplus activation, the rank and the set of adapted projections are unchanged, and only the recurrent cell differs; parameter counts stay within $5\%$ of the Mamba modulator ($+2.4\%$ for GRU, $+4.9\%$ for LSTM). At the segment level the frozen encoder slice, attention pool, recurrence width and scoring heads are unchanged, and a two-layer GRU or LSTM replaces the two-layer Mamba.

\begin{table}[!h]
  \centering\small
  \setlength{\tabcolsep}{3pt}
  \caption{\textbf{Token level: recurrent family in the MaLoRA modulator.} F1, mean $\pm$ standard deviation over three seeds. The three families perform similarly, with no consistent winner.}
  \label{tab:recurrent_token}
  \begin{tabular}{l cc cc}
    \toprule
    & \multicolumn{2}{c}{MuSiQue} & \multicolumn{2}{c}{2WikiMultihopQA} \\
    \cmidrule(lr){2-3}\cmidrule(lr){4-5}
    Family & Qwen & Llama & Qwen & Llama \\
    \midrule
    GRU   & $56.5${\scriptsize$\pm1.6$} & $64.3${\scriptsize$\pm0.9$} & $78.6${\scriptsize$\pm0.8$} & $83.6${\scriptsize$\pm0.5$} \\
    LSTM  & $57.1${\scriptsize$\pm0.4$} & $62.6${\scriptsize$\pm0.6$} & $79.8${\scriptsize$\pm0.2$} & $83.4${\scriptsize$\pm0.3$} \\
    Mamba & $56.4${\scriptsize$\pm1.1$} & $62.3${\scriptsize$\pm1.8$} & $79.3${\scriptsize$\pm0.5$} & $83.3${\scriptsize$\pm0.7$} \\
    \bottomrule
  \end{tabular}
\end{table}

\begin{table}[!h]
  \centering\small
  \setlength{\tabcolsep}{3pt}
  \caption{\textbf{Segment level: recurrent family in the MaRA mixer.} Recall@$4$, no adapter attached, matched skeleton (Section~\ref{sec:routing-ablation}). The selective state-space module leads every cell.}
  \label{tab:recurrent_segment}
  \begin{tabular}{l cc cc}
    \toprule
    & \multicolumn{2}{c}{MuSiQue} & \multicolumn{2}{c}{2WikiMultihopQA} \\
    \cmidrule(lr){2-3}\cmidrule(lr){4-5}
    Family & Qwen & Llama & Qwen & Llama \\
    \midrule
    GRU   & $60.2$ & $58.7$ & $94.8$ & $93.8$ \\
    LSTM  & $58.6$ & $58.6$ & $94.9$ & $93.5$ \\
    Mamba & $\mathbf{80.6}$ & $\mathbf{84.1}$ & $\mathbf{99.2}$ & $\mathbf{99.5}$ \\
    \bottomrule
  \end{tabular}
\end{table}

The two levels give opposite answers. At the token level (Table~\ref{tab:recurrent_token}) no family consistently dominates: four-cell means are $70.8$ for GRU, $70.7$ for LSTM and $70.3$ for Mamba, and the per-cell differences are comparable to seed-level variability. The largest gap, $+2.0$ F1 for GRU on MuSiQue Llama, is of the same order as the seed standard deviations it sits between ($0.9$ and $1.8$). What separates the token-level modulator from the stateless baseline of Table~\ref{tab:modulation_main} is therefore recurrence itself, not the selective parameterisation.

At the segment level (Table~\ref{tab:recurrent_segment}) the same substitution costs $20$ to $25$ points of Recall@$4$ on MuSiQue. The segment sequence is short and each position summarises a whole paragraph, so the selective state-space parameterisation captures behaviour that the matched GRU and LSTM alternatives do not reproduce.

Training wall-clock is within $0.4\%$ across the three families at the token level ($56.6$ks for GRU against $56.4$ks for Mamba on MuSiQue Qwen), so no efficiency claim is made for the selective formulation at the scales used here; the advantage claimed is the asymptotic one of a parallel scan.

\section{Smaller backbone: Qwen-2.5-3B}
\label{app:small_backbone}

The token-level comparison is repeated on Qwen-2.5-3B, roughly half the parameter count of the smallest backbone in the main grid. Training and evaluation follow Section~\ref{sec:modulation-results}: rank $r{=}16$, epoch-$2$ checkpoints, beam $4$, and the full validation split ($2417$ MuSiQue, $3000$ 2WikiMultihopQA). One seed per cell is run rather than three.

\begin{table}[!h]
  \centering\small
  \setlength{\tabcolsep}{2.0pt}
  \caption{\textbf{Token modulators on Qwen-2.5-3B} (EM / F1, single seed).}
  \label{tab:small_backbone}
  \begin{tabular}{l cc}
    \toprule
    Method & MuSiQue & 2WikiMultihopQA \\
    \midrule
    LoRA \citep{hu2022lora}       & $37.3/49.3$ & $70.0/75.8$ \\
    TopLoRA \citep{li2025beyond}  & $35.4/47.9$ & $64.3/70.4$ \\
    MaLoRA                        & $\mathbf{38.7/51.1}$ & $\mathbf{72.3/77.8}$ \\
    \bottomrule
  \end{tabular}
\end{table}

MaLoRA leads on both datasets, by $+1.8$ F1 on MuSiQue and $+2.0$ F1 on 2WikiMultihopQA over LoRA, and by $+3.2$ and $+7.4$ F1 over TopLoRA. The ordering of MaLoRA above LoRA is preserved from the $7$--$9$B grid of Table~\ref{tab:modulation_main}. TopLoRA falls below LoRA at this scale on both datasets, whereas in the main grid it trails LoRA only on Gemma-2-9B MuSiQue. Because a single seed is run per cell, differences below roughly $2$ F1 should not be read as reliable; the per-seed standard deviations at $7$--$9$B reach $1.8$ F1 (Table~\ref{tab:seed_std}).

MaRA is also trained at this scale, under the protocol of Table~\ref{tab:router_arch}: encoder depth $K{=}16$, single scoring pass, one seed, $n{=}500$ validation examples. Table~\ref{tab:small_backbone_mara} reports the result against the two untrained retrieval baselines measured on the same backbone. On 2WikiMultihopQA the $3$B backbone matches the $7$B one ($99.1$ against $99.2$). On MuSiQue recall falls from $80.6$ to $72.0$. Retrieval quality therefore tracks backbone capacity on the harder selection problem, as expected of a component that reads the backbone's own representations. The margin over the untrained baselines is preserved at both scales.

\begin{table}[!h]
  \centering\small
  \setlength{\tabcolsep}{4pt}
  \caption{\textbf{MaRA on Qwen-2.5-3B} (Recall@$4$, single seed, single scoring pass, $K{=}16$). Frozen-LM cosine and MaRA are measured on the same $3$B backbone. BM25 \citep{Robertson2009ThePR} is backbone-independent and is repeated from Table~\ref{tab:retrieval_compare}. MuSiQue frozen-LM cosine uses $n{=}498$; all other cells use $n{=}500$.}
  \label{tab:small_backbone_mara}
  \begin{tabular}{l cc}
    \toprule
    Method & MuSiQue & 2Wiki \\
    \midrule
    BM25             & $47.0$ & $66.4$ \\
    Frozen-LM cosine & $40.4$ & $49.5$ \\
    MaRA             & $\mathbf{72.0}$ & $\mathbf{99.1}$ \\
    \bottomrule
  \end{tabular}
\end{table}

\section{Commonsense reasoning}
\label{app:commonsense}

Commonsense reasoning is a complementary single-context testbed: each example is a self-contained multiple-choice question, with no candidate paragraphs and no multi-hop chain to retrieve. MaRA does not apply and is omitted; only the token-level modulators are compared to LoRA. Adapters are trained on the standard $10{,}000$-example Commonsense suite \citep{hu-etal-2023-llm} and use the same rank $r{=}16$ and schedule as the main experiments. Table~\ref{tab:commonsense} reports the mean accuracy over eight multiple-choice tasks (BoolQ, PIQA, SIQA, HellaSwag, WinoGrande, ARC-easy, ARC-challenge, OpenBookQA).

\begin{table}[!h]
  \centering\small
  \setlength{\tabcolsep}{2pt}
  \caption{\textbf{Commonsense accuracy} (\%, mean over eight multiple-choice tasks). Single-context; MaRA omitted. Best non-base per column in bold.}
  \label{tab:commonsense}
  \begin{tabular}{l ccc}
    \toprule
    Method & Qwen & Llama & Gemma \\
    \midrule
    Base    & $77.1$ & $67.6$ & $68.7$ \\
    LoRA    & $88.3$ & $80.6$ & $\mathbf{86.6}$ \\
    TopLoRA & $86.8$ & $79.7$ & $84.5$ \\
    MaLoRA  & $\mathbf{88.5}$ & $\mathbf{83.3}$ & $86.2$ \\
    \bottomrule
  \end{tabular}
\end{table}

On Llama-3.1-8B MaLoRA improves over LoRA ($+2.7$) while TopLoRA is slightly below ($-0.9$); on Qwen-2.5-7B MaLoRA is level with LoRA ($+0.2$) and TopLoRA is below ($-1.5$); on Gemma-2-9B all three fine-tuned variants are tightly clustered (LoRA $86.6$, TopLoRA $84.5$, MaLoRA $86.2$; all $+15.8$ to $+17.9$ over the $68.7$ base). Across backbones MaLoRA wins on average ($86.0$ vs.\ LoRA $85.2$ vs.\ TopLoRA $83.7$). Commonsense reasoning is largely saturated for these backbones once fine-tuned, so per-cell differences are small relative to the multi-hop gains in Table~\ref{tab:main} and consistent with little headroom for token-level reshaping in the absence of multi-paragraph distractor structure.

\section{Additional evaluations}
\label{app:additional}

Four evaluations that fall outside the main grid are reported here. All reuse checkpoints, ranks and training schedules already described. No hyperparameter was retuned for them.

\subsection{Mathematical reasoning}

Table~\ref{tab:math} reports $5$-shot exact match on GSM8K \citep{cobbe2021gsm8k} and GSM-hard \citep{pmlr-v202-gao23f}, scored on the full $1{,}319$-example test split under greedy decoding. Evaluation is on Qwen-2.5-7B. The rank, module set, optimiser and schedule match Table~\ref{tab:main}. Only token-level modulation is evaluated. Both tasks present a single reasoning context with no paragraph structure for MaRA to select over.

MaLoRA improves over LoRA by $+0.9$ EM on GSM8K and by $+2.1$ EM on GSM-hard. The margin is smaller than the $+6.4$ F1 mean observed on multi-hop question answering. This is consistent with the account in Section~\ref{sec:results}. A single reasoning context offers a shorter trajectory over which recurrent state can accumulate. The result is reported as preliminary evidence beyond retrieval-centred question answering rather than as a general claim for mathematical reasoning.

\begin{table}[!h]
  \centering\small
  \caption{\textbf{Mathematical reasoning} ($5$-shot exact match, Qwen-2.5-7B, full $1{,}319$-example test split, greedy decoding). Rank, module set and training schedule match Table~\ref{tab:main}.}
  \label{tab:math}
  \begin{tabular}{lcc}
    \toprule
    Method & GSM8K & GSM-hard \\
    \midrule
    LoRA \citep{hu2022lora} & $77.4$ & $42.3$ \\
    MaLoRA                  & $\mathbf{78.3}$ & $\mathbf{44.4}$ \\
    \bottomrule
  \end{tabular}
\end{table}

\subsection{Rank-matched low-rank baseline}

The modulator adds parameters over the LoRA baseline, so a rank-matched control is required to separate the mechanism from the budget. LoRA is retrained at rank $r{=}24$ with $\alpha/r{=}2$ under the schedule of Appendix~\ref{app:hparams}, on MuSiQue with Qwen-2.5-7B and Llama-3.1-8B, at one seed. At this rank the LoRA baseline carries the same trainable-parameter count as AdaLoRA on both backbones, and comes within $5\%$ of MaLoRA.

Extra rank does not close the gap. Rank $24$ scores below rank $16$ on both backbones, by $3.1$ and $5.8$ F1, so the schedule that suits rank $16$ does not transfer to the larger budget. MaLoRA exceeds the rank-matched baseline by $8.4$ and $13.7$ F1 while adding roughly $5\%$ more parameters than it. The improvement is therefore not explained by parameter count.

\begin{table}[!h]
  \centering\small
  \setlength{\tabcolsep}{3pt}
  \caption{\textbf{Rank-matched comparison} (MuSiQue F1). Trainable parameters are given per backbone. LoRA at rank $24$ is parameter-matched to AdaLoRA on both backbones. The rank-$24$ row is a single seed; the other rows follow the seed protocol of Appendix~\ref{app:hparams}.}
  \label{tab:rank_matched}
  \begin{tabular}{l cc cc}
    \toprule
    & \multicolumn{2}{c}{Params (M)} & \multicolumn{2}{c}{MuSiQue F1} \\
    \cmidrule(lr){2-3}\cmidrule(lr){4-5}
    Method & Qw & Ll & Qw & Ll \\
    \midrule
    LoRA, $r{=}16$ & $27.1$ & $28.3$ & $51.1$ & $54.4$ \\
    LoRA, $r{=}24$ & $40.6$ & $42.5$ & $48.0$ & $48.6$ \\
    AdaLoRA        & $40.6$ & $42.5$ & $53.9$ & $59.7$ \\
    MaLoRA         & $42.5$ & $44.6$ & $\mathbf{56.4}$ & $\mathbf{62.3}$ \\
    \bottomrule
  \end{tabular}
\end{table}

\subsection{Retrieval beyond the training datasets}

Two checks address whether MaRA is specific to the datasets on which it was trained. Both report Recall@$4$ on $n{=}500$ held-out examples with gold supporting-paragraph labels, matching the protocol of Table~\ref{tab:retrieval_compare}.

MaRA was first trained and evaluated on HotpotQA \citep{yang2018hotpotqa}, a third multi-hop benchmark carrying supporting-paragraph labels. Recall@$4$ reaches $97.9$ on both Qwen-2.5-7B and Llama-3.1-8B, at the per-backbone encoder depth selected on validation ($K{=}16$ and $K{=}12$ respectively).

A MuSiQue-trained MaRA was then applied to 2WikiMultihopQA with no further training. Recall@$4$ reaches $85.9$ on Qwen-2.5-7B and $89.7$ on Llama-3.1-8B. Both exceed every untrained retriever in Table~\ref{tab:retrieval_compare}, including the $8$B-parameter Qwen3-Embedding encoder at $83.5$. The read-out therefore transfers across multi-hop datasets rather than fitting the one on which it was trained.

\section{MaRA architecture details}
\label{app:router_details}

This appendix expands the condensed Section~\ref{sec:routing} with per-stage discussion, the architecture-ablation deep dive, and the full classical-retrieval comparison.

\subsection{Per-stage notes}

\paragraph{Segment encoder.}
The frozen encoder is the same backbone used downstream, truncated to its first $K$ layers. Each segment $C_i$ is concatenated with the query $Q$ before encoding, so the resulting hidden states are already query-conditioned. No new encoder parameters are introduced. Encoding each (query, segment) pair independently is the standard cross-encoder pattern.

\paragraph{$Q$-driven attention pool.}
The attention pool of Eq.~\ref{eq:pool} reduces each segment's $T_i$-step hidden-state sequence to a single embedding $p_i$ via additive attention with a learned query vector $q_a$. An ablation replacing the learned attention pool by a plain mean is reported in Table~\ref{tab:router_arch} as MeanPool.

\paragraph{Mamba over segments.}
The two-layer selective state-space module \citep{gu2024mamba} of Eq.~\ref{eq:mamba_seg} follows the same SSM update as Eq.~\ref{eq:ssm}, but the recurrence runs over \emph{segments} rather than tokens. This is the key architectural choice and is ablated against TokenMamba (token-level Mamba) and Transformer (segment-level transformer) in Table~\ref{tab:router_arch}.

\paragraph{Scoring heads.}
Eq.~\ref{eq:score} uses two linear heads: a local head $\mathbf{w}_{\ell}$ that scores each segment from its own recurrent state $s_i$ (used for top-$k$ selection), and a global head $W_g$ that reads the final state $g = s_N$ and predicts all $N$ supporting labels jointly as an auxiliary signal. There is no per-segment MLP combiner; global context reaches each score through the causal recurrence (each $s_i$ depends on segments $1$ through $i$) and through the shared parameters that the global-head objective conditions. The recurrence's contribution is isolated by the PoolOnly ablation (mixer removed) and by the MLP ablation (per-segment MLP without recurrence) in Table~\ref{tab:router_arch}.

\paragraph{Auxiliary supervision.}
The global-head term in Eq.~\ref{eq:loss-router} ($\gamma{=}0.5$) is an auxiliary BCE over all $N$ paragraphs, predicted jointly from the final state $g = s_N$. It shapes the shared recurrence but does not feed selection, and its contribution to final recall is small.

\paragraph{Composition with the modulator.}
MaRA operates before the modulated language model's forward pass; the modulator (Section~\ref{sec:modulation}) operates inside it. The two share only the language model's hidden states as a representation substrate. Training is staged: the modulated adapter is trained first on the target task; MaRA is then trained against the unadapted backbone, with no modulator attached.

\subsection{Architecture ablation (full analysis)}

The Mamba over segments wins on every cell of Table~\ref{tab:router_arch}. The R@$4$ gap to the second-best architecture (MeanPool, which keeps segment-level Mamba but drops the attention pool) is $7$~pp on both Qwen and Llama for MuSiQue. TokenMamba uses the same Mamba module as MaRA but applied to the concatenated token sequence; its R@$4$ is $19$ to $29$~pp below the segment-level Mamba on MuSiQue, isolating the contribution of placing the recurrence at the segment level. The encoder follows the standard cross-encoder pattern, encoding each (query, segment) pair independently, with a trained Mamba mixer on top of the per-segment embeddings carrying the cross-segment interaction \citep{karpukhin-etal-2020-dense, nogueira2020passagererankingbert}. On 2WikiMultihopQA, where the candidate set is smaller and the supporting evidence is less distractor-rich, every architecture except LastTok and TokenMamba reaches at least $90\%$ R@$4$ and the differences between architectures are compressed, but the Mamba still leads on every cell.

\subsection{Scoring-head ablation: linear vs.\ MLP}
The local scoring head (Eq.~\ref{eq:score}) is a single linear map $\mathbf{w}_\ell$. Replacing it with a two-layer MLP ($\mathrm{Linear}\!\to\!\mathrm{GELU}\!\to\!\mathrm{Linear}$, hidden $256$), with the Mamba mixer and all other settings fixed, does not help: Recall@$4$ is $78.6$ vs.\ $80.6$ (MuSiQue Qwen), $83.2$ vs.\ $84.1$ (MuSiQue Llama), $99.4$ vs.\ $99.2$ (2WikiMultihopQA Qwen), and $99.4$ vs.\ $99.5$ (2WikiMultihopQA Llama). The MLP head is marginally worse on MuSiQue and level on 2WikiMultihopQA. The recurrent state $s_i$ already integrates cross-segment context, so a linear read-out suffices; the extra head capacity adds parameters without improving retrieval. The linear head is used throughout.

\subsection{Retrieval baselines (full comparison)}

MaRA is compared against a $2{\times}2$ grid of alternatives, crossing \emph{outsourced} (a separate encoder over raw text) versus \emph{model-attached} (reusing the frozen backbone's own hidden states), and \emph{untrained} (off-the-shelf or cosine similarity) versus \emph{trained} (fine-tuned on the supporting-paragraph supervision). Table~\ref{tab:retrieval_compare} reports Recall@$4$ on the same $n{=}500$ validation split and gold labels for every method.

\begin{table*}[!h]
  \centering\small
  \setlength{\tabcolsep}{8pt}
  \caption{\textbf{Retrieval comparison} (Recall@$4$, $n{=}500$ validation, identical examples and gold labels). Outsourced methods run a separate encoder over raw text; model-attached methods score from the frozen backbone's own hidden states. Trained = fine-tuned on the supporting-paragraph labels. Best per column in bold. MaRA beats every outsourced baseline, including the $8$B Qwen3-Embedding encoder, while adding only $\approx 3$M parameters and no separate encoder.}
  \label{tab:retrieval_compare}
  \begin{tabular}{lcc}
    \toprule
    Method & MuSiQue & 2WikiMultihopQA \\
    \midrule
    \multicolumn{3}{l}{\textit{Outsourced, off-the-shelf (untrained)}} \\
    BM25 (sparse) \citep{Robertson2009ThePR}       & $47.0$ & $66.4$ \\
    E5-large-v2 \citep{wang2024textembeddingsweaklysupervisedcontrastive}                & $69.1$ & $83.7$ \\
    Qwen3-Embedding-8B \citep{zhang2025qwen3embeddingadvancingtext}     & $75.5$ & $83.5$ \\
    \midrule
    \multicolumn{3}{l}{\textit{Outsourced, trained (dense bi-encoder, best of LR sweep)}} \\
    E5-large-v2 (fine-tuned) \citep{wang2024textembeddingsweaklysupervisedcontrastive}          & $63.3$ & $97.1$ \\
    Contriever (fine-tuned) \citep{izacard2022unsuperviseddenseinformationretrieval} & $68.3$ & $96.9$ \\
    BGE-large-en (fine-tuned) \citep{10.1145/3626772.3657878}        & $68.2$ & $97.7$ \\
    \midrule
    \multicolumn{3}{l}{\textit{Model-attached, untrained (frozen-LM cosine)}} \\
    Frozen-LM cosine (Qwen-2.5-7B)                & $37.7$ & $54.9$ \\
    Frozen-LM cosine (Llama-3.1-8B)               & $48.4$ & $48.2$ \\
    \midrule
    \multicolumn{3}{l}{\textit{Model-attached, trained (ours)}} \\
    MaRA (Qwen-2.5-7B)                            & $80.6$ & $99.2$ \\
    MaRA (Llama-3.1-8B)                           & $\mathbf{84.1}$ & $\mathbf{99.5}$ \\
    \bottomrule
  \end{tabular}
\end{table*}

Three readings. First, MaRA wins every cell: on MuSiQue it reaches $80.6$ (Qwen) and $84.1$ (Llama), above the strongest outsourced baseline, the $8$B Qwen3-Embedding encoder ($75.5$), despite using no separate encoder and only $\approx 3$M new parameters. Second, training a \emph{separate} dense bi-encoder on the supporting-paragraph labels is not enough: across three encoders (E5, Contriever, BGE-large-en), fine-tuning lifts 2WikiMultihopQA to $96.9$--$97.7$ but plateaus on MuSiQue at $63.3$--$68.3$, well below MaRA ($80.6$--$84.1$). For E5 the MuSiQue fine-tune even \emph{degrades} relative to its own zero-shot ($69.1 \to 63.3$), a catastrophic-forgetting effect when a general-purpose encoder is specialised on a small label set. The MuSiQue ceiling for text-only retrieval reflects that the bridge entity is absent from the surface query, which a separate encoder cannot recover but the backbone's reasoning hidden states can. Third, the untrained model-attached baseline (raw frozen-LM cosine) is weak ($37.7$--$54.9$), so the gain is not simply from reusing the backbone's hidden states; it comes from the trained segment-level mixer on top of them. BM25 R@$8$/R@$12$ reach $60.8$/$71.5$ (MuSiQue) and $86.2$/$100$ (2WikiMultihopQA); Qwen3-Embedding R@$8$/R@$12$ reach $88.9$/$93.9$ and $96.4$/$100$.

\subsection{Base against adapted hidden states}
\label{app:base_vs_adapted}

The MaRA encoder reads hidden states from the frozen backbone. Those states can be taken from the unmodified base model, or from a backbone already carrying a trained adapter. Table~\ref{tab:base_vs_adapted} compares four sources under one matched campaign. Encoder depth, candidate pool, schedule, seed and scoring protocol match Table~\ref{tab:router_arch}. Only the source of the hidden states changes. The base condition reproduces the corresponding Table~\ref{tab:router_arch} entry in all four cells.

Supporting-paragraph recall is largely insensitive to the choice. The spread across the four sources is at most $2.2$ points, and falls to $0.4$ on 2WikiMultihopQA, where every condition exceeds $99$. Adapted states are strongest on both MuSiQue cells, so adaptation is mildly helpful there rather than neutral. The effect stays small in context. The same adapters change end-task F1 on the corresponding cells by more than $36$ points (Table~\ref{tab:main}), an order of magnitude more than they change recall. Evidence relevance is therefore already recoverable from the unmodified base representations. The retrieval adapter mainly decodes information the frozen model already carries.

\begin{table}[!h]
  \centering\small
  \setlength{\tabcolsep}{2.0pt}
  \caption{\textbf{Base against adapted hidden states} (MaRA supporting-paragraph Recall@$4$, $n{=}500$ validation, single seed). The encoder reads the unmodified base model, or a backbone carrying a trained LoRA, TopLoRA or MaLoRA adapter. All other settings match Table~\ref{tab:router_arch}.}
  \label{tab:base_vs_adapted}
  \begin{tabular}{l ccccc}
    \toprule
    Cell & Base & LoRA & TopLoRA & MaLoRA & Spread \\
    \midrule
    MuSiQue Qw & $80.6$ & $81.6$ & $82.6$ & $82.8$ & $2.2$ \\
    MuSiQue Ll & $84.1$ & $83.6$ & $84.3$ & $85.3$ & $1.7$ \\
    2Wiki Qw   & $99.2$ & $99.4$ & $99.2$ & $99.2$ & $0.2$ \\
    2Wiki Ll   & $99.5$ & $99.4$ & $99.8$ & $99.4$ & $0.4$ \\
    \bottomrule
  \end{tabular}
\end{table}

\section{Chain segment selection}
\label{app:iter}
MaRA (Section~\ref{sec:routing}) is applied over $N_p$ chained passes at inference. Given query $Q$, candidate segments $\mathcal{C}=\{C_1,\dots,C_N\}$, a pass budget $N_p$, and a selection budget $k$, an ordered anchor set is built one segment per pass; each pass conditions the segment scoring on the anchors already chosen (Algorithm~\ref{alg:iter}). $N_p{=}4$ is used on both datasets, matching the pass budget used to train the retrieval adapter. During training, the intermediate passes $1{:}N_p{-}1$ receive an auxiliary loss weight and the final pass weight $1.0$.

\begin{algorithm}[!h]
\caption{Chain segment selection}
\label{alg:iter}
\begin{algorithmic}[1]
\STATE \textbf{Input:} query $Q$, segments $\mathcal{C}=\{C_1,\dots,C_N\}$, passes $N_p$, budget $k$, adapter $R$
\STATE \textbf{Output:} ordered segment indices, truncated to $k$
\STATE $A \gets [\,]$ \quad\COMMENT{ordered anchor list}
\FOR{$t = 1$ \TO $N_p$}
  \STATE $\mathbf{r} \gets R\big(\mathcal{C}, Q \mid A\big)$ \quad\COMMENT{score each segment, conditioned on selected anchors}
  \STATE $a_t \gets \arg\max_{i \notin A}\, r_i$
  \STATE $A \gets A \mathbin{\|} [\,a_t\,]$
\ENDFOR
\STATE $\mathrm{rest} \gets \mathrm{sort}\big(\{i : i \notin A\}\big)$ by $r_i$ descending \quad\COMMENT{$\mathbf{r}$ from the final pass}
\STATE \textbf{return} $(A \mathbin{\|} \mathrm{rest})[:k]$
\end{algorithmic}
\end{algorithm}

\section{Top-$k$ selection sweep}
\label{app:per-cell}

\begin{table}[!t]
  \centering\small
  \setlength{\tabcolsep}{4pt}
  \caption{\textbf{MaRA supporting-paragraph recall@$k$ by backbone}: mean fraction of gold supporting paragraphs retained, over three seeds. Bold marks the $k$ used throughout.}
  \label{tab:recall_per_bb}
  \begin{tabular}{l ccc c}
    \toprule
    & \multicolumn{3}{c}{MuSiQue R@$k$} & 2Wiki R@$k$ \\
    \cmidrule(lr){2-4}\cmidrule(lr){5-5}
    Backbone & $k{=}4$ & $k{=}8$ & $\mathbf{k{=}12}$ & $\mathbf{k{=}4}$ \\
    \midrule
    Qwen   & $0.86$ & $0.96$ & $\mathbf{0.98}$ & $\mathbf{1.00}$ \\
    Llama  & $0.86$ & $0.95$ & $\mathbf{0.98}$ & $\mathbf{1.00}$ \\
    Gemma  & $0.70$ & $0.87$ & $\mathbf{0.94}$ & $\mathbf{0.99}$ \\
    \bottomrule
  \end{tabular}
\end{table}

Per-backbone MaRA recall@$k$ is given in Table~\ref{tab:recall_per_bb}, measured on the same runs that produce Table~\ref{tab:main}. The rule that fixes $k$ per dataset, and the recall figures behind it, are given in Section~\ref{sec:results}. End-task F1 confirms the choice. Re-running MaLoRA$+$MaRA on MuSiQue at $k{=}8$ changes F1 by $+0.7$ (Qwen), $-0.5$ (Llama) and $-3.3$ (Gemma), a mean of $-1.0$, with the largest loss on the backbone whose recall is weakest.

\end{document}